\pdfoutput=1
\documentclass[11pt]{article}

\usepackage{ACL2023}

\usepackage{times}
\usepackage{latexsym}

\usepackage[T1]{fontenc}
\usepackage[utf8]{inputenc}

\usepackage{microtype}

\usepackage{inconsolata}

\usepackage{graphicx}
\graphicspath{{images/}}

\usepackage{multirow}
\usepackage{multicol}
\usepackage{tabularx}
\usepackage{varwidth}
\usepackage{adjustbox,lipsum}
\usepackage{booktabs}

\usepackage{color,soul}
\usepackage{times}
\usepackage{latexsym}
\usepackage{graphicx}
\usepackage{pgfplots}
\usepackage{booktabs}
\usepackage{multirow}
\usepackage{comment}
\usepackage{changepage}
\usepackage{colortbl}
\usepackage{textcomp}
\usepackage{subcaption}
\usepackage{graphicx}

\pgfplotsset{compat=1.18}

\PassOptionsToPackage{dvipsnames}{xcolor}

\newcommand{\secref}[1]{\S\ref{#1}} % Define a new command, refer to section with "§"

\newcommand{\dataset}{{\sc CovidET-EXT}}

\newcommand{\CovidET}{{\sc CovidET}}

\usepackage[textsize=scriptsize]{todonotes}

\newcommand{\red}[1]{\textcolor{purple}{#1}}
\newcommand{\blue}[1]{\textcolor{blue}{#1}}

\sethlcolor{lime}
\newcommand{\gray}[1]{\textcolor{lightgray}{#1}}
\title{Unsupervised Extractive Summarization of Emotion Triggers}

 \author{
    \textbf{Tiberiu Sosea$^*$}$^1$\quad\textbf{Hongli Zhan$^*$}$^2$\quad\textbf{Junyi Jessy Li}$^2$\quad\textbf{Cornelia Caragea}$^1$\\
    $^1$Department of Computer Science, University of Illinois Chicago\\$^2$Department of Linguistics, The University of Texas at Austin\\
    {\color{blue}\texttt{\{tsosea2,cornelia\}@uic.edu}\quad\texttt{\{honglizhan,jessy\}@utexas.edu}}
}

\begin{document}
\maketitle

\begingroup\begin{NoHyper}\def\thefootnote{*}\footnotetext{Tiberiu Sosea and Hongli Zhan contributed equally.}\end{NoHyper}\endgroup

%\cornelia{alternate title: Unsupervised Emotion-Trigger  Summarization}

\begin{abstract}
Understanding what leads to emotions during large-scale crises is important as it can provide groundings for expressed emotions and subsequently improve the understanding of ongoing disasters. Recent approaches \cite{zhan-etal-2022-feel} trained supervised models to both detect emotions and explain emotion triggers (events and appraisals) via abstractive summarization. However, obtaining timely and qualitative abstractive summaries is expensive and extremely time-consuming, requiring highly-trained expert annotators. In time-sensitive, high-stake contexts, this can block necessary responses. We instead pursue \emph{unsupervised} systems that extract triggers from text. First, we introduce \dataset{}, augmenting \cite{zhan-etal-2022-feel}'s abstractive dataset (in the context of the COVID-19 crisis) with extractive triggers. Second, we develop new unsupervised learning models that can jointly detect emotions and summarize their triggers. Our best approach, entitled Emotion-Aware Pagerank, incorporates emotion information from external sources combined with a language understanding module, and outperforms strong baselines. We release our data and code at \url{https://github.com/tsosea2/CovidET-EXT}.
%by an average of $1.1\%$ in ROUGE-L.
\end{abstract}

\section{Introduction}
Language plays a central role in social, clinical, and cognitive psychology \cite{Pennebaker-2003}, and social media presents a gold mine for such analysis: people turn to social media to share experiences around challenges in their personal lives and seek diagnosis, treatment, and emotional support for their conditions \cite{Choudhury2014MentalHD, gjurkovic-snajder-2018-reddit}. During crises, such as natural disasters or global pandemics, large-scale analysis of language on social media --- both \emph{how people feel} and \emph{what's going on in their lives to lead to these feelings} --- can have a profound impact on improving mental health solutions as well as helping policymakers take better-informed decisions during a crisis.

\begin{figure}[t]
    \centering
    \includegraphics[width=\columnwidth]{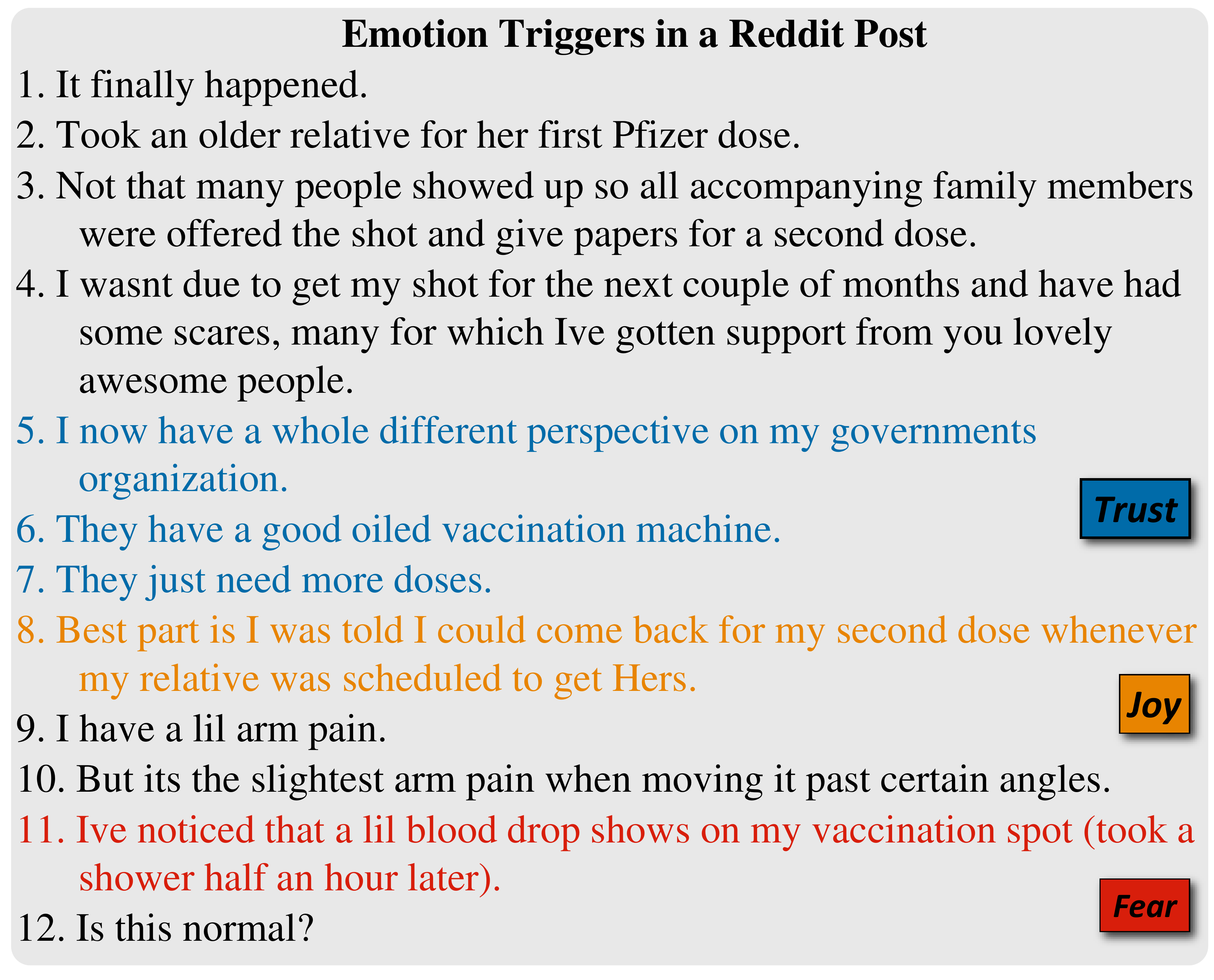}
    \caption{An example post from \dataset{} annotated with emotion triggers. The highlighted sentences represent triggers of the tagged emotions.}
    \label{fig:example}
\end{figure}

%\looseness=-1
Recent work \cite{zhan-etal-2022-feel} 
taps into this broad challenge by jointly detecting emotions and generating a natural language description about what triggers them (triggers include both objective events and subjective appraisals of those events~\cite{ellsworth2003appraisal,moors-appraisal-2013}). Trigger explanation is formulated as a supervised, abstractive summarization task that is emotion-specific. %\jl{I meant brags for THIS dataset, not CovidET! Can you add to the description of OUR dataset?}\hz{done on line 82} 
Unlike generic summarization however, due to the high cognitive load to provide judgments \emph{for each emotion}, obtaining human-written summaries for this task is time-consuming and requires significant annotator training. This results in small, domain-specific datasets that are difficult to scale --- especially in the face of new crisis events where the timing of such analysis is often pivotal.

This work instead takes a fully \emph{unsupervised} approach such that we do not rely on any labeled data, thus becoming agnostic to distributional shifts in domain or types of crisis, and robust for time-critical events. We posit that emotion triggers can be summarized effectively in an extractive manner where unsupervised methods are well-suited; we thus tackle the challenge of \emph{simultaneous} emotion prediction and trigger extraction.

For this new task, we first introduce \dataset{}, augmenting \citet{zhan-etal-2022-feel}'s {\sc CovidET} with
manually annotated extractive summaries corresponding to each of their abstractive summaries. The result is
a dataset of $1,883$ Reddit posts about the COVID-19 pandemic, manually annotated with $7$ fine-grained emotions (from {\sc CovidET}) and their corresponding \textbf{extractive} triggers (Figure \ref{fig:example}). For every emotion present in a post, our annotators highlight sentences that summarize the emotion triggers, resulting in $6,741$ extractive summaries in total. Qualitative analyses of the dataset indicate good agreement among the annotators, and follow-up human validations of the annotations also reveal high correctness.
% \hz{done} 
\dataset{} provides an ideal test bed to facilitate the development of extractive (supervised or unsupervised) techniques for the tasks of emotion detection and trigger summarization in crisis contexts.

\looseness=-1
We propose Emotion-Aware PageRank (EAP), a novel, fully unsupervised, graph-based approach for extractive emotion trigger summarization from text. The core of our method is to decompose the traditional PageRank \cite{Page1999ThePC} ranking algorithm into multiple biased PageRanks \cite{1208999}, one for each emotion. To bias our model towards various emotions, our approach harnesses lexical information from emotion lexicons \cite{mohammad2013crowdsourcing,mohammad-2018-word}. Critically, unlike previous graph-based unsupervised approaches \cite{mihalcea-tarau-2004-textrank,liu-etal-2010-automatic,ctr,florescu-caragea-2017-positionrank,patel-caragea-2021-exploiting,singh2019embedding}, which represent the text as a bag-of-words or word embeddings, EAP incorporates a language understanding module leveraging large language models to ensure that the summaries for an emotion are coherent in the context of that emotion. Results on our \dataset{} indicate the effectiveness of our EAP, which significantly pushes the Rouge-L score of our summaries by an average of $2.7\%$ over strong baselines.

% We\jl{This needs significant rework. Tiberu will edit this.} carry out extensive experiments using several unsupervised extractive summarization approaches to establish a baseline on our dataset. To push the performance of our baselines,
% We propose a new approach which we call Emotion-Aware Pagerank (\ourmethod). Our algorithm is built on top of TextRank \cite{mihalcea-tarau-2004-textrank}, a graph-based, text-ranking algorithm. EAP integrates two powerful components that enable its application in our emotion context. First, we leverage external information from emotion lexicons \cite{mohammad-2018-word} to enable our model to pay greater attention to lexical cues associated with emotions. Second, we incorporate a highly-effective language understanding module that extracted summaries \emph{make sense} in the larger context of the detected emotion.

Our contributions are as follows: \textbf{1)} We introduce \dataset{}, a manually annotated benchmark dataset for the task of emotion detection and trigger summarization. \textbf{2)} We propose Emotion-Aware PageRank, a variation of PageRank that combines a language understanding module and external emotion knowledge to generate emotion-specific extractive summaries. \textbf{3)} We carry out a comprehensive set of experiments using numerous baselines to evaluate the performance on \dataset{} and show that our proposed EAP significantly outperforms strong baselines.

\section{Background and Related Work}

\paragraph{Emotion Tasks.} Most of the prior work on emotions on social media focuses solely on detecting emotions or emotional support from text \cite{Wang2012HarnessingT, biyani-etal-2014-identifying, abdul-mageed-ungar-2017-emonet, Khanpour_Caragea_Biyani_2018, khanpour-caragea-2018-fine,demszky-etal-2020-goemotions, desai-etal-2020-detecting,sosea-caragea-2020-canceremo,adikari_achini,calbi_m,KABIR2021100135,beck-etal-2021-investigating,10.1145/3308560.3316599,sosea-caragea-2021-emlm,hosseini-caragea-2021-distilling-knowledge,Hosseini_Caragea_2021,saakyan-etal-2021-covid,ils-etal-2021-changes,sosea-etal-2022-emotion,sosea-caragea-2022-ensynet,sosea-caragea-2022-leveraging}. Our task is directly related to emotion cause extraction~\cite{GAO20154517, gui-etal-2016-event, Gao2017OverviewON} which focused on identifying phrase-level causes from Chinese news or micro-blogs, which are distinct from the spontaneous writing on social media. In our context, similar to the work of \citet{zhan-etal-2022-feel}, what \emph{triggers} an emotion includes both what happened and how the writer appraised the situation. A major difference of our work from \citet{zhan-etal-2022-feel} is that we consider extractive summaries instead of abstractive and take a fully unsupervised perspective, eliminating the reliance on labeled data. For a comprehensive overview of \CovidET{} introduced by \citet{zhan-etal-2022-feel}, refer to Appendix \secref{covid-et}.
% Additionally, prior work in emotion cause extraction developed phrase-level corpora in Chinese news (distinct from spontaneous writing on social media) or Chinese micro-blogs (very short text). They also rely on explicit emotion keywords, rather than human perceived emotions.
% \jl{@Hongli please merge+shorten these two paragraphs!}

% \jl{No, I don't mean CovidET, I mean you need to talk about how OUR dataset in THIS paper is good in the paragraph describing CovidET!}

% \jl{@Tibi this paragraph needs work, please make a pass.}
%\noindent
\paragraph{Unsupervised Extractive Summarization.}
Extractive summarization aims to condense a piece of text by identifying and extracting a small number of important sentences \citep{Allahyari-2017-review, liu-lapata-2019-text, El-Kassas-2021-survey} that preserve the text's original meaning. The most popular approaches in unsupervised extractive summarization leverage graph-based approaches to compute a sentence's salience for inclusion in a summary \cite{mihalcea-tarau-2004-textrank,zheng-lapata-2019-sentence}. These methods represent sentences in a document as nodes in an undirected graph whose edges are weighted using sentence similarity. The sentences in the graph are scored and ranked using node centrality, computed recursively using PageRank \cite{Page1999ThePC}. In contrast, our EAP considers words instead of sentences as nodes in the graph and employs multiple separate biased PageRanks \cite{1208999} to compute an emotion-specific score for each word, which is combined with a sentence-similarity module to produce one sentence score per emotion, indicating the salience of the sentences under each emotion. 

% Among the various types of automatic text summarization systems, query-based summarization and aspect-based summarization are two similar sub-tasks to our emotion-based trigger summarization. Query-based summarization produces summaries from documents that are most relevant to a given search query \cite{zhong-etal-2021-qmsum, El-Kassas-2021-survey}. On the other hand, the objective of aspect-based summarization is to generate summaries with respect to different aspects or perspectives \cite{tan-etal-2020-summarizing, hayashi-etal-2021-wikiasp}.

% Drawing from query-based and aspect-based summarization techniques, we propose the emotion-based trigger summarization approach to address the current task of Emotion-Trigger Summarization. Different from the above techniques, we perform trigger summarization with regard to automatically detected emotions in the document. 

\section{Dataset Construction}
Since there is no annotated data for extractive emotion triggers summarization in crisis contexts, we first bridge this gap by extending \CovidET{}, \citet{zhan-etal-2022-feel}'s abstractive-only dataset with extractive trigger summaries. Doing so \textbf{(a)} creates benchmark data for extractive systems; \textbf{(b)} allows in-depth analyses to understand how and when emotion triggers are expressed on social media. This will also create a parallel abstractive-extractive dataset for future research. We name our new dataset \dataset{} ({\sc CovidET} \{extractive, extension\}).

\paragraph{Annotating Emotion Triggers.}
Given a post from \CovidET{} annotated with an emotion $e$, we ask annotators to highlight sentences in the post that best describe the trigger for $e$. An overview of our annotation scheme can be viewed in Appendix \secref{appendix:annotation-scheme}. We recruit both undergraduate students (in a Linguistics department) as well as pre-qualified crowd workers (from the Amazon Mechanical Turk) for this task.\footnote{These crowd workers have an ongoing working relationship with our group and have prior experience in related complex tasks, and we make sure they are paid at least \$$10$/hr.} %Both groups of annotators were trained in a \textit{qualification} process \ts{reference?example?description?}. 
Each post is annotated by two annotators. %\jl{I don't get what you are saying here, neither text nor comment...}
We monitor the annotation quality and work with the annotators during the full process. Similar to {\sc CovidET}, the test set is annotated by undergraduate students.

\paragraph{Benchmark Dataset.} We follow the benchmark setup in \citet{zhan-etal-2022-feel} with $1,200$ examples for training, $285$ examples for validation, and $398$ examples for testing. If two annotators highlight different sentences as triggers for the same emotion, we consider both sets of sentences as the gold summaries and evaluate them using multi-reference ROUGE. We anonymize \dataset{}. Note that since we explore {\em unsupervised} methods, the training set is {\em not} used in our summarization models. Nevertheless, we emphasize that while the focus of this work is the unsupervised setup, we hope that \dataset{} can spur further research into both supervised and unsupervised methods, hence we maintain the splits in \citet{zhan-etal-2022-feel}. For completeness, we carry out experiments in a fully supervised setup in Appendix \secref{supervised-extractive-summarization}.

% \subsection{Validation and Agreement}

% \ts{please check, jessy's comment commented out here -> want to assess space }
% \hz{done}
% \jessy{Please shrink/merge \% overlap and metric overlap (can just be stated in a table), add Kappa. Also move human validation to be the first thing that is presented, (maybe?) move abstractive-extractive alignment here too}\hz{Re-written}

%\looseness=-1
%\noindent
\paragraph{Human Validation.} We validate the annotated extractive summaries of emotion triggers in \dataset{} through inspections from third-party validators on the Amazon Mechanical Turk crowdsourcing platform. A subset of our training data including $300$ randomly selected examples which contain annotations of extractive summaries of emotion triggers are validated. Given an annotated extractive trigger summary, we first ask the validators whether the summary leans towards the annotated emotion. It yes, we ask the validator to further point out if the \textit{trigger} --- rather than the \textit{emotion} itself --- is present in the summary. The percentage of examples that validators confirm for the two steps is shown in Table \ref{tab:dataset-validation-ext}. Overall, the human validation results showcase moderately high correctness in the annotations of \dataset{}, considering the subjective nature of our task.\footnote{The same sentence can be interpreted to be triggers for different emotions. For example, the sentence ``I miss my room and I dont have many clothes or my meds here, but hes hitting these mics every fucking night and Im scared of contracting it'' expresses \textit{anger}, \textit{sadness}, and \textit{fear} simultaneously under the same context.}

\begin{table}[t]
    \centering
    \adjustbox{max width=\linewidth}{
    \begin{tabular}{l|ccccccccc}
        \toprule
        % Table generated by Excel2LaTeX
        & \textbf{ANC} & \textbf{AGR} & \textbf{FER}  & \textbf{SDN} & \textbf{JOY}   & \textbf{TRS} & \textbf{DSG} & \textbf{Avg} \\
        \midrule
        \textbf{Emotion} & $0.64$  & $0.84$  & $0.84$  & $0.84$  & $0.92$  & $0.60$  & $0.80$  & $0.79$ \\
        \textbf{Trigger} & $0.56$  & $0.64$  & $0.76$  & $0.76$  & $0.80$  & $0.56$  & $0.72$  & $0.69$ \\
        \bottomrule
    \end{tabular}
    }
    \caption{Human validation results on \dataset{}.} %Scores indicate the percentage of times the emotion / trigger is present in the summary.}
    \label{tab:dataset-validation-ext}
\end{table}

\setlength\extrarowheight{2pt}
\begin{table}[t]
    \centering
    \small
    \rowcolors{2}{gray!25}{white}
    \begin{tabularx}{\columnwidth}{| >{\hsize=.4\hsize}X | >{\hsize=.6\hsize}X|}
        \rowcolor{gray!25}
        \hline
        \textbf{Overlapping Status} & {$55.5$\% of all summaries}\\\hline
        \textbf{Fleiss' Kappa}   &   {$0.89$ across $7$ emotions}\\\hline
        \textbf{self-BLEU-2}    &   {$0.429$ (baseline: $0.151$)}\\\hline
        \textbf{self-BLEU-3}    &   {$0.419$ (baseline: $0.139$)}\\\hline
        \textbf{self-ROUGE-L}    &   {$0.504$ (baseline: $0.229$)}\\\hline
    \end{tabularx}
    \caption{Inter-annotator statistics of \dataset{}.}
    \label{tab:overlap-statistics}
\end{table}

\noindent
\paragraph{Inter-Annotator Agreement.} We measure the inter-annotator agreement between two extractive trigger summaries for the same emotion in a post, as shown in Table \ref{tab:overlap-statistics}. Results show that, within the examples where we find emotion overlaps, $29.9$\% of the extractive summaries of triggers for the same emotion share completely identical annotations from both annotators, and $25.6$\% have partial sentence-level overlaps. In total, we find overlaps in $55.5$\% of the summaries, and the experts who were responsible for the test set ($65.8$\%) have more overlapping summaries than the crowd workers who were responsible for the training and validation sets ($52.3$\%). Furthermore, the average Fleiss' kappa \cite{fleiss-1971, randolph2005free} is $0.89$ across all the emotions in \dataset{}. This suggests substantial agreement among our annotators.%\hz{done}

In addition, we also employ automatic metrics including self-BLEU (with smoothing methods $1$) and self-ROUGE to capture the overlap between annotators' summaries. To establish a baseline, we report these metrics between the annotators' work and a randomly selected sentence from the original post. We repeat this process five times. Results reveal that both the self-BLEU and self-ROUGE of our annotations significantly outperform that of the random baseline (as shown in Table \ref{tab:overlap-statistics}). %Moreover, the performance of the undergraduate students consistently exceeds the turkers in terms of these diversity metrics 
We also observed higher values of these measures for student annotators compared with crowd workers.
(c.f. Appendix \secref{agreement}). These results indicate strong accordance among our annotators.

\begin{figure}[t]
    \centering
    \includegraphics[width=\columnwidth]{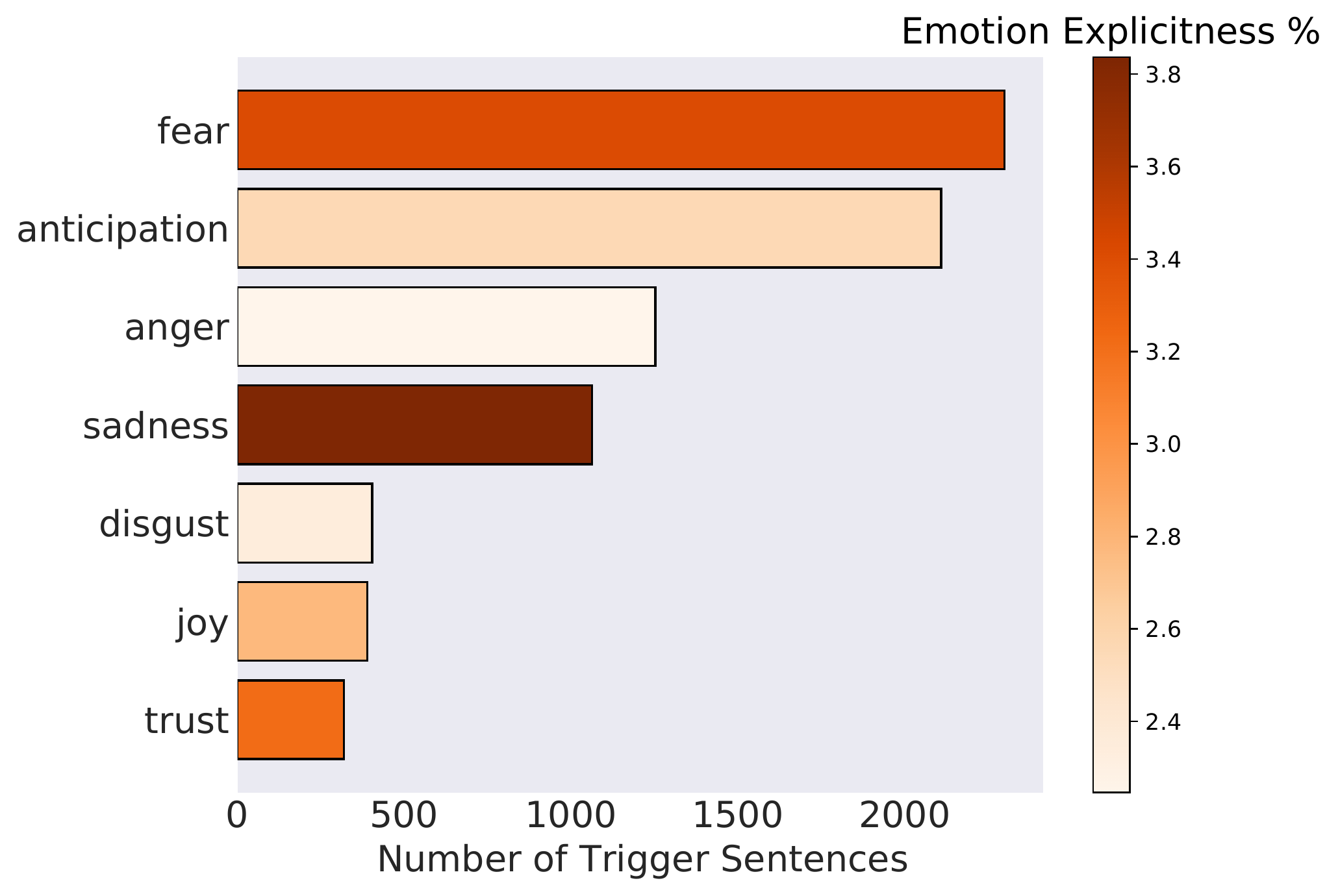}
    \caption{The sentence-level distribution of triggers in the original posts in \dataset{}. The colorbar shows the explicitness of the emotion in the triggers.}
    \label{fig:ext-emolex}
\end{figure}

\paragraph{Dataset Statistics.} Here we elaborate on the overview of \dataset{}. On average, there are $1.35$ sentences (std.dev $= 0.79$) consisting of $32.54$ tokens (std.dev $= 20.68$) per extractive summary of emotion trigger in \dataset{}. As shown in Figure \ref{fig:ext-emolex}, when broken down into unique trigger \textit{sentences}, \textit{fear} has the most trigger sentences in the dataset, closely followed by \textit{anticipation}. On the other hand, \textit{trust} has the lowest number of trigger sentences. This can be attributed to the calamitous nature of the domain of our dataset. Besides, unlike generic news summarization~\cite{fabbri2021summeval}, the emotion-trigger extractive summarization task is \emph{not} lead-based. This is manifested through our scrutiny of the position of emotion trigger sentences in the original posts (Figure \ref {fig:ext-trigger-position} and Figure \ref{fig:ext-trigger-heatmap}, Appendix \secref{appendix:data-analysis}), where a large number of triggers cluster in the later parts of the post.

Additional analyses of \dataset{} can be found in Appendix \secref{appendix:data-analysis}.%\hz{done}

\noindent
\paragraph{Emotion Explicitness.} To examine the explicitness of emotions in the extractive summaries of emotion triggers, we apply EmoLex \cite{mohammad2013crowdsourcing}, an English lexicon for the Plutchik-8 primary emotions. Specifically, for the extractive summaries of triggers to a certain emotion $e$, we measure the average ratio of $e$'s words in EmoLex being present in the sentence-level lemmatized summaries. The results are presented in Figure \ref{fig:ext-emolex}. Interestingly, we notice that \textit{sadness} is the most explicit emotion in the annotated extractive summaries of triggers in our dataset, while \textit{anger} is the most implicit one.
% \jl{There should be a direct connection to modeling. Can you check results -- is sadness the easiest to detect? And is anger the hardest?}

\section{Unsupervised Extractive Summarization}
\begin{figure*}[t]
    \centering
    \resizebox{\linewidth}{!}{
    \includegraphics[width=\textwidth]{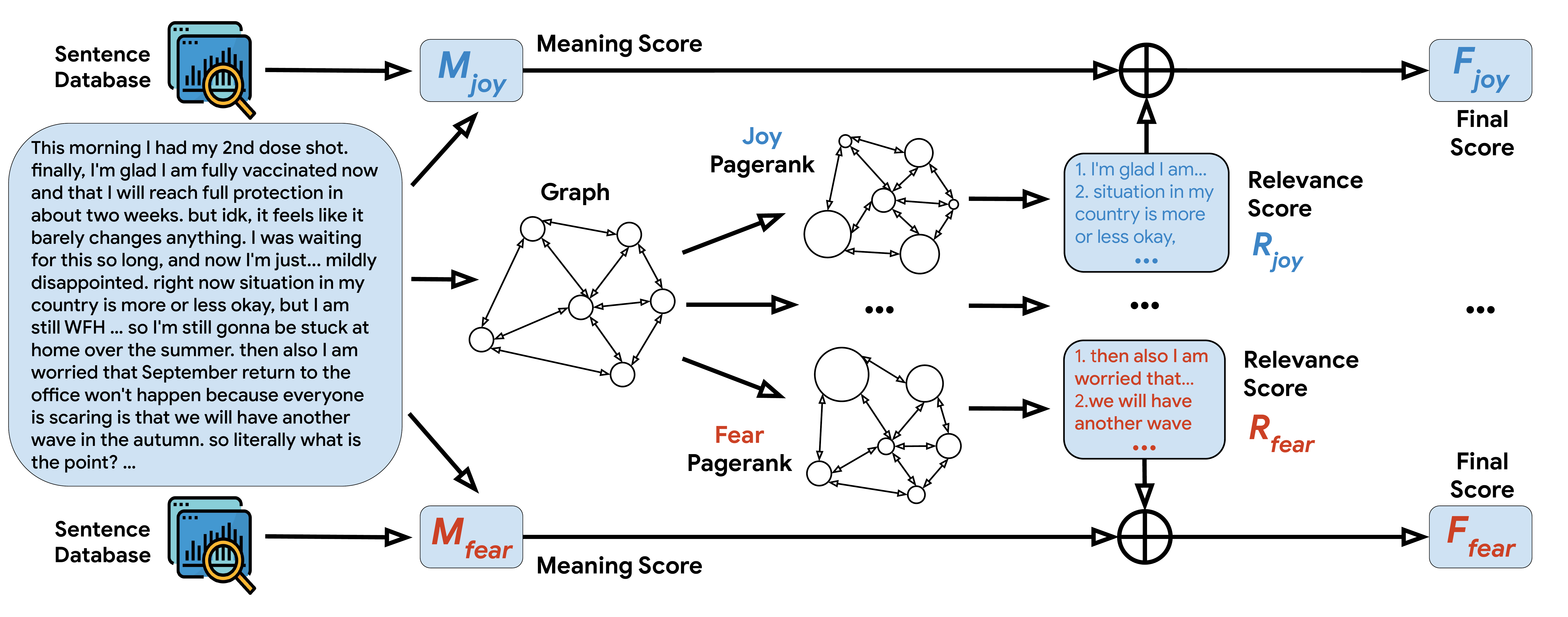}
    }
    \caption{Diagram of our Emotion-Aware PageRank. EAP builds a word graph from a post, then runs separate biased PageRanks, one for each emotion, to score every candidate sentence under each emotion. The score is combined with an emotion-aware language understanding module to produce final rankings for each sentence under each emotion.}
    \label{fig:model architecture}
    \vspace{-4mm}
\end{figure*}

In this section we introduce Emotion-Aware Pagerank (EAP), our fully unsupervised, graph-based, emotion trigger extractive summarization method that incorporates information from emotion lexicons to calculate a biased PageRank score of each sentence in a post. EAP then fuses this score with an additional similarity-based sentence-level score that ensures the summary for a specific emotion $e$ does not diverge in meaning from other summaries of the same emotion $e$. We show an overview of our model architecture in Figure \ref{fig:model architecture}.
% \jl{diverge from what? Don't get it...}

%\noindent
\paragraph{Task Formulation.} Let $P$ be a Reddit post. $P$ is composed of an ordered sequence of $n$ sentences: $P = \{s_{1}, s_{2}, ..., s_{n}\}$. Generic extractive summarization aims to output an ordered set of sentences $S$ with $S \subset P$ that captures the essence of post $P$. In our emotion trigger summarization, however, we aim to generate multiple extractive summaries conditioned on the expressed emotions. To this end, we are interested in a set of summaries $S^{emo} = \{S_{e_{1}}, S_{e_{2}}, ..., S_{e_{m}}\}$ where $m$ is the total number of emotions present in $P$ and $S_{e_{i}}$ is the summary of the triggers that lead to the expression of emotion $e_{i}$ with $S_{e_{i}} \subset P$. Note that $P$ usually conveys a subset of emotions, in which case the summaries for the emotions that are not present in text are empty.

% \jl{What happens if an emotion is not there? empty summary? better make it clear.}
\paragraph{Graph Construction.} We build an undirected graph $G=(V, E)$, where $V$ is vocabulary set of words. To build $V$ we employ various processing and filtering techniques. First, we only select nouns, adjectives, verbs, adverbs and pronouns and remove any punctuation. Next, we stem all the selected words to collapse them in a common base form. Finally, we remove infrequent words which appear less than $20$ times in the entire training set. The remaining words form the vocabulary $V$. A pair of words $(w_{i}, w_{j}) \in E$ defines an edge between $w_{i}$ and $w_{j}$ and the operator $\beta(w_{i}, w_{j})$ denotes the weight of edge $(w_{i}, w_{j})$. We compute the weight of an edge in our graph using word co-occurences in windows of text. Given a window size of $ws$, we say that two words $w_{i}$ and $w_{j}$ co-occur together if the number of words between them in text is less than $ws$. We build a co-occurence matrix $C$ of size $|V|\times|V|$ from the documents in our training set where $C_{ij}$ is the number of times words $w_{i}$ and $w_{j}$ co-occur together. Using $C$ we simply define the weight of an edge as:

\vspace{-7mm}
\begin{equation}
    \beta(w_{i}, w_{j}) = \frac{2 \times C_{ij}}{\sum_{k=0}^{|V|}(C_{ik} + C_{jk})}
\end{equation}
\vspace{-3mm}

\noindent
Intuitively, the more frequently two words co-occur together, the higher the weight of the edge between them becomes.

% and show detailed information about the window size as well as its effect on the model performance in Appendix .\jl{missing ref}

\paragraph{Emotion Decomposition.} In PageRank, the importance or relevance $\mathcal{R}(w_{i})$ of an arbitrary word $w_{i}$ is computed in an iterative fashion using the following formula:

\vspace{-6.15mm}
\begin{equation}
    \mathcal{R}(w_{i}) = \lambda \sum_{k = 1}^{|V|}  \beta(w_{k}, w_{i})\mathcal{R}(w_{k}) + (1-\lambda)\frac{1}{|V|}
\label{simple-pagerank-importance}
\end{equation}
\vspace{-5.8mm}

\noindent
where $|.|$ is the set size operator and $\lambda$ is the damping factor, a fixed value from $0$ to $1$ which measures the probability of performing a random jump to any other vertex in the graph. The idea of PageRank is that a vertex or word is important if other important vertices point to it. The constant term $\frac{1}{|V|}$ is called a random jump probability and can be viewed as a node \emph{preference} value, which in this case assigns equal weights to all the words in the graph, indicating no preference.

In this current formulation, the PageRank model calculates the weights of words irrespective of the expressed emotion. We claim that for our purpose words should bear different importance scores in different emotion contexts. For example, the word \emph{agony} should have a higher importance in the context of \emph{sadness} or \emph{fear} than in the context of \emph{joy}.

To this end, we propose to decompose the text into multiple components, one for each emotion, where the relevance of a word differs from component to component. Biased PageRank \cite{1208999} is a variation of PageRank where the second term in Equation \ref{simple-pagerank-importance} is set to be non-uniform, which can influence the algorithm to prefer particular words over others. We propose to run a separate biased PageRank for each emotion and leverage a custom importance function $i_{e}(w_{i})$ that yields high values for words that are correlated with an emotion $e$ and low values otherwise. Formally, the relevance computation for the PageRank corresponding to emotion $e$ becomes:

% traditional PageRank algorithm into multiple biased PageRanks \cite{1208999}, one for each emotion. Biased PageRank \cite{1208999} is a variation of PageRank where the second term in Equation \ref{simple-pagerank-importance} (i.e., the \emph{preference} value) is set to be non-uniform. In this case, the algorithm will prefer those words with high \emph{preference}. Given an emotion $e$, we bias our PageRank by replacing this term with an importance function $i_{e}(w_{i})$ that yields high values for words that are correlated with emotion $e$ and low values otherwise. Formally,

\vspace{-3mm}
\begin{equation}
    \mathcal{R}_{e}(w_{i}) = \lambda \sum_{k = 1}^{|V|}  \beta(w_{k}, w_{i})\mathcal{R}_{e}(w_{k}) + (1-\lambda)\frac{i_{e}(w_{i})}{N}
\label{emotion_aware_pagerank}
\end{equation}
\vspace{-3mm}

\noindent
where $N$ is a normalization factor such that $\sum_{w \in V}\frac{i_{e}(w)}{N} = 1$. Since the model prefers those vertices with higher random jump probabilies, using an accurate importance function $i_{e}(w_{i})$ for emotion $e$ can lead to accurate relevance scores in the context of $e$.  We define this function using the NRC emotion intensity \cite{mohammad-2018-word} lexicon. EmoIntensity associates words with their expressed emotions and also indicates the degree of correlation between a word and a particular emotion using real values from $0$ to $1$. For example, \emph{outraged} has an intensity for anger of $0.964$ while \emph{irritation} has an intensity of $0.438$. In our context, assigning importance values using intensity is appropriate since a sentence containing high intensity words for an emotion $e$ is more likely to be relevant in the context of $e$ compared to a sentence containing lower intensity words. Denoting the set of words in EmoIntensity correlated with emotion $e$ by $\mathcal{I}_{e}$, all words $w \in \mathcal{I}_{e}$ also come with intensity value annotations denoted by $int_{e}(w)$. Therefore, we define the importance function as:

\vspace{-2mm}
\begin{equation}
i_{e}(w)= \left\{
\begin{array}{ll}
      int_{e}(w) \quad if \quad w  \in \mathcal{I}_{e}\\
      c \quad if \quad w \in V \setminus  \mathcal{I}_{e} \\
\end{array} 
\right.
\label{importance_function}
\end{equation}

\noindent
where $c$ is a constant that we find using the validation set. Since our summaries are at the sentence level, we simply score a sentence $s_{i}$ as the average relevance of its words:

\begin{equation}
    R_{e}(s_{i}) = \frac{\sum_{w_{j} \in s_{i}} R_{e}(w_{j})}{|s_{i}|}
\label{sentence_importance}
\end{equation}

\paragraph{Encoding the meaning.} A major drawback of prior graph-based approaches is that they exclusively represent the input as a bag-of-words,
% \jl{I aasdssume you do not mean the BOW representations but rather there is no global context? Can you make that clearer?} \ts{I mean BOW, i.e., the order does not matter} 
ignoring the structure of text. We propose to solve this drawback by introducing a language model-based component to encode the meaning of a sentence. Our component is based on the assumption that a sentence $s$ that is highly relevant for an emotion $e$ should be similar in meaning to other sentences $s_{i}$ relevant to $e$. We capture this property by scoring each sentence based on its similarity with other important (i.e., in the context of $e$) sentences. We leverage the popular Sentence-BERT \cite{reimers-gurevych-2019-sentence} model, which produces meaningful sentence embeddings that can be used in operations such as cosine similarity. Given a sentence $s_{i}$, let $\mathbf{s_{i}}$ be its embedding and $sim(\mathbf{s_{i}}, \mathbf{s_{j}})$ be the cosine similarity between the embeddings of sentences $s_{i}$ and $s_{j}$. Denoting by $\mathcal{T}$ the set of sentences in the entire dataset, we score $s_{i}$ in the context of emotion $e$ as follows:

\begin{equation}
    M_{e}(s_{i}) = \frac{\sum_{s \in \mathcal{T}}^{} sim(\mathbf{s_{i}}, \mathbf{s}) * \mathcal{R}_{e}(s)}{|\mathcal{T}|}
\end{equation}

\noindent
Intuitively, $M_{e}(s_{i})$ yields high values if $s_{i}$ is similar in meaning to sentences relevant in the context of emotion $e$.

\paragraph{Constructing the Summaries.} Given a post $P = \{s_{1}, s_{2},...,s_{n}\}$, we first combine the meaning and the relevance scores into a final, sentence level, per-emotion score, which we use to score every sentence $s_{i}$ in $P$ along all the emotions:

\begin{equation}
    \mathcal{F}_{e}(s_{i}) = \mathcal{R}_{e}(s_{i}) * M_{e}(s_{i})
\label{final_threshold}
\end{equation}

\noindent
\looseness=-1
We use this per-emotion score to rank the sentences in the post $P$. For an emotion $e$, we only select the sentences $s_{i}$ where $\mathcal{F}_{e}(s_{i}) > t$ to be part of the final summary for $e$. $t$ is a threshold value that we infer using our validation set. Note that given $P$, we compute the score $\mathcal{F}_{e}$ for every emotion $e$. In the case that none of the sentences in $P$ exceed the threshold for a particular emotion, we consider that the emotion is not present in the post (i.e., we do not generate a summary).

% none of the sentences in a post might exceed this threshold for some emotions. In this case, we consider that these emotions are not present in the post.

\begin{table*}[!htbp]
\setlength{\tabcolsep}{3pt}
\centering
\small
\resizebox{16cm}{!}{%
\begin{tabular}{r|cc|cc|cc|cc|cc|cc|cc|cc}
 & \multicolumn{2}{c}{\textsc{anger}} & \multicolumn{2}{c}{\textsc{disgust}} & \multicolumn{2}{c}{\textsc{fear}} & \multicolumn{2}{c}{\textsc{joy}} & \multicolumn{2}{c}{\textsc{sadness}} & \multicolumn{2}{c}{\textsc{trust}} & \multicolumn{2}{c}{\textsc{anticipation}} & \multicolumn{2}{c}{\textsc{avg}} \\
& R-2 & R-L & R-2 & R-L & R-2 & R-L & R-2 & R-L & R-2 & R-L & R-2 & R-L & R-2 & R-L & R-2 & R-L \\
 \toprule
\textsc{1-sent}	& $0.174$ & $0.240$	& $0.095$	& $0.170$ &  $0.202$	& 	$0.256$	& $0.119$ &	$0.179$ & $0.110$ &	$0.177$ & $0.189$ &$0.236$ & $0.160$ & $0.220$& $0.149$ &	$0.211$ \\ 
\textsc{3-sent}	& $0.301$ & $0.315$	& $0.196$ &$0.253 $ & $0.322$ & $0.343$	& $0.273$ & 
 $0.310$& $0.239$ &	$0.292$ & $0.248$ & $0.279$& $0.263$ &	$0.307$ & $0.258$ &	$0.288$ \\ 
\textsc{PacSum}	& $0.308$ & $0.314$ & $0.210$ &$0.218$	& $0.327$ &	$0.331$& $0.276$ &	$0.282$& $0.287$ &$0.304$	& $0.225$	&$0.234$	& $0.283$  &	$0.295$	& $0.273$ & $0.282$ \\ 
\textsc{PreSumm} & $0.306$ & $0.312$& $0.219$ &	$0.221$& $0.332$ &	$0.335$& $0.268$ &	$0.274$	& $0.295$ &$0.317$	& $0.222$ &	$0.227$	& $0.284$ &	$0.291$	& $0.275$ &	$0.282$	 \\
\textsc{TextRank}	& $0.296$ & $0.301$& $0.236$ &$0.235$	& $0.319$ &	$0.326$& 
$0.272$ & $0.276$	& $0.286$ &$0.306$& $0.225$ &$0.231$	& $0.218$ 	&$0.221$	&	$0.264$&$0.270$ \\
\midrule
\textsc{EmoLex}	& $0.213$ & $0.260$	& $0.218$ &$0.256$	&	$0.309$ &$0.341$ &  $0.218$ &$0.252$& $0.301$ &$0.331$& $0.176$ & $0.203$	& $0.207$ &	$0.242$& $0.234$ &	$0.269$ \\
\textsc{EmoIntensity}& $0.307$	& $0.322$ & $0.269$ &$0.281$ & $0.342$ &$0.355$	& $0.222$ & $0.235$ & $0.329$ & $0.341$	& $0.227$	& $0.242$	& $0.295$ 	&$0.310$	& $0.284$ &	$0.298$	 \\
\textsc{BERT-GoEmo} & $0.247$ &	$0.264$ & $0.232$ &	$0.237$ & $0.296$ &	$0.312$	& $0.221$ & $0.247$	& $0.314$ & $0.321$	& $0.201$ &	$0.204$	& $0.247$ &$0.225$	& 
 $0.253$ &$0.258$	 \\
\midrule
\textsc{EAP} & $\mathbf{0.324^{\dagger}}$  &	$\mathbf{0.348^{\dagger}}$	& $\mathbf{0.285^{\dagger}}$ &	$\mathbf{0.296^{\dagger}}$	& $\mathbf{0.364^{\dagger}}$ &$\mathbf{0.373^{\dagger}}$& $\mathbf{0.285^{\dagger}}$ & $\mathbf{0.319^{\dagger}}$& $\mathbf{0.348^{\dagger}}$ &	$\mathbf{0.354^{\dagger}}$&	 $\mathbf{0.258^{\dagger}}$ &$\mathbf{0.291^{\dagger}}$	& $\mathbf{0.319^{\dagger}}$ &$\mathbf{0.324^{\dagger}}$	& $\mathbf{0.309^{\dagger}}$ &$\mathbf{0.325^{\dagger}}$	 \\
\bottomrule
\end{tabular}
}
\caption{Results of our models in terms of ROUGE-2 and ROUGE-L. We assert significance$^{\dagger}$ using a bootstrap test where we resample our dataset $50$ times with replacement (with a sample size of $500$) and $p<0.05$.}
\label{tab:main_table}
\vspace{-2mm}
\end{table*}

\section{Experiments and Results}
In this section, we first introduce our emotion-agnostic and emotion-specific baselines. Next, we present our experimental setup and discuss the results obtained by EAP against the baselines.

\paragraph{Emotion-agnostic baselines.} We explore two standard heuristic baselines, namely \textbf{1)} Extracting the first sentence in the post ($1$ sent) and \textbf{2)} Extracting the first three sentences in the post ($3$ sent). Next, we design three graph centrality measure-based methods: \textbf{3)} PacSum \cite{zheng-lapata-2019-sentence}, \textbf{4)} PreSum \cite{liu-lapata-2019-text} and word-level \textbf{5)} TextRank \cite{mihalcea-tarau-2004-textrank}.
Note that these methods are emotion-oblivious and the generated summary will be identical for different emotions. 

\begin{table}[!htbp]
\setlength{\tabcolsep}{3pt}
\centering
\small
\adjustbox{max width = \columnwidth}{
\begin{tabular}{r|cccccccc}
 & \textsc{ang} & \textsc{dsg} & \textsc{fer} & \textsc{joy} & \textsc{sdn} & \textsc{trt} & \textsc{anc} & \textsc{avg} \\
 \toprule
{\small \textsc{EmoLex}}  & $0.561$ & $0.572$ & $0.568$  & $0.613$ & $0.563$ & $0.581$ & $0.593$ &  $0.578$ \\
{\small \textsc{EmoIntensity}}  & $0.581$ & $0.583$ & $0.557$ & $0.632$ & $0.573$ & $0.589$ & $0.585$ & $0.584$ \\
{\small \textsc{GoEmotions}} & $0.516$  & $0.532$ & $0.562$ & $0.576$ & $0.531$ & $0.556$ & $0.574$ & $0.537$\\ 
% \textsc{HEm} & $0.507$ & $0.514$ & $0.548$ & $0.554$ & $0.496$ & $0.527$ & $0.573$ & $0.527$ \\ 
% \textsc{CEm} & $0.523$ & $0.521$ & $0.528$ & $0.569$ & $0.520$ & $0.567$ & $0.582$ & $0.527$ \\
\midrule
{\small \textsc{EAP}} & $\mathbf{0.593^{\dagger}}$  & $\mathbf{0.595^{\dagger}}$ & $\mathbf{0.583}$ & $\mathbf{0.649^{\dagger}}$ &  $\mathbf{0.581^{\dagger}}$  & $\mathbf{0.606^{\dagger}}$  & $\mathbf{0.612^{\dagger}}$ &  $\mathbf{0.593^{\dagger}}$ \\
\bottomrule
\end{tabular}
}
\caption{Emotion detection results of our models in terms of Macro F-1. We assert significance$^{\dagger}$ using a bootstrap test where we resample our dataset $50$ times with replacement (with a sample size of $500$) and $p<0.05$.}
\label{tab:emotion-detection}
\vspace{-2mm}
\end{table}

\paragraph{Emotion-specific baselines.}
We first employ two lexical-based methods: \textbf{6)} EmoLex - we use the EmoLex \cite{mohammad2013crowdsourcing} lexicon to identify lexical cues that indicate the expression of emotions. If a sentence contains a word that is associated with an emotion $e$, we consider the sentence to express $e$. The final summary for $e$ contains all sentences expressing $e$. \textbf{7)} EmoIntensity - we leverage the NRC Affect Intensity Lexicon \cite{mohammad-2018-word} to build a more fine-grained approach of identifying if a sentence expresses an emotion or not. For each sentence and emotion, we calculate the average emotion word intensity and compare it to a pre-defined threshold $t$. If the average intensity for $e$ is higher than $t$ we label the sentence with $e$. $t$ is a tunable parameter that we select based on our validation set performance.

% \jl{??? These are just emotion recognition models doesn't mean we are doing DA at all! This is super confusing and inaccurate, please take this out. Also, why do we need all these datasets for emotions in the first place? I think we should move results for hurricaneemo and canceremo to appendix.} 

Finally, we leverage models trained on emotion detection datasets to build our emotion-specific summaries.  For a post $P$, we use our model to make predictions on each sentence in $P$ and build summaries by concatenating sentences that express the same emotions. We mainly experiment with a model trained on the \textbf{8)} GoEmotions \cite{demszky-etal-2020-goemotions} dataset.

% how well the information translates between domains and how well it applies to our extractive summarization context. For a post $P$, we use a model to make predictions on each sentence in $P$. After obtaining emotion labels for each sentence, we perform a post-processing step to build summaries by concatenating sentences that are labeled with the same emotion

% , maintaining the relative order of sentences from $P$. We train a BERT base uncased model \cite{devlin-etal-2019-bert} on three datasets from various domains, yielding three approaches:  \textbf{5)} GoEmotions \cite{demszky-etal-2020-goemotions}, a dataset of general Reddit comments; \textbf{6)} CancerEmo \cite{sosea-caragea-2020-canceremo}, which is composed of online health-related posts; and \textbf{7)} HurricaneEmo \cite{desai-etal-2020-detecting} of tweets during hurricanes. 

% Next, 

% \paragraph{Strong Baselines} First, we experiment with two emotion-oblivious methods, PacSum \cite{zheng-lapata-2019-sentence} and PreSumm \cite{liu-lapata-2019-text}. Additionally, we design two methods based on surface-level lexical information: \textbf{1)}  \textbf{2)} 

% \cornelia{why are EmoLex and EmoIntensity strong baselines? we should organize them differently - 1/3 first sentences in one block; emolex/emointensity in another block; BERT based with DA in another block; graph-based centrality measures in another block - PacSum, PreSumm;PageRank; we could also compare with PositionRank; and/or if you wish TPR.}

\begin{table*}[t]
\setlength{\tabcolsep}{3pt}
\centering
\small
\resizebox{16cm}{!}{%
\begin{tabular}{r|cc|cc|cc|cc|cc|cc|cc|cc}
 & \multicolumn{2}{c}{\textsc{anger}} & \multicolumn{2}{c}{\textsc{disgust}} & \multicolumn{2}{c}{\textsc{fear}} & \multicolumn{2}{c}{\textsc{joy}} & \multicolumn{2}{c}{\textsc{sadness}} & \multicolumn{2}{c}{\textsc{trust}} & \multicolumn{2}{c}{\textsc{anticipation}} & \multicolumn{2}{c}{\textsc{avg}} \\
& R-2 & R-L & R-2 & R-L & R-2 & R-L & R-2 & R-L & R-2 & R-L & R-2 & R-L & R-2 & R-L & R-2 & R-L \\
\toprule
\textsc{EAP} & $\mathbf{0.324}$  &	$\mathbf{0.348}$	& $\mathbf{0.285}$ &	$\mathbf{0.296}$	& $\mathbf{0.364}$ &$\mathbf{0.373}$& $\mathbf{0.285}$ & $\mathbf{0.268}$& $\mathbf{0.348}$ &	$\mathbf{0.354}$&	 $\mathbf{0.239}$ &$\mathbf{0.264}$	& $\mathbf{0.319}$ &$\mathbf{0.324}$	& $\mathbf{0.309}$ &$\mathbf{0.318}$	 \\
\midrule
-int & $0.317$ &$0.336$ &$0.274$ &$0.282$ &$0.353$ &$0.362$&$0.276$&$0.261$&$0.339$&$0.347$&$0.231$&$0.252$&$0.312$&$0.317$&$0.300$&$0.308$ \\
-sim & $0.314$ &$0.332$ &$0.277$ &$0.284$ &$0.351$ &$0.360$&$0.272$&$0.260$&$0.340$&$0.342$ & $0.232$&$0.254$&$0.311$&$0.31$& $0.299$ &$0.306$ \\
-int 
-sim& $0.300$ &$0.316$ &$0.263$ &$0.275$ &$0.341$ &$0.353$&$0.261$&$0.253$&$0.325$&$0.339$&$0.224$&$0.247$&$0.308$& $0.309$&$0.28$&$0.298$ \\
\bottomrule
\end{tabular}
}
\caption{Ablation study of our EAP.}
\label{tab:ablation}
\end{table*}

\begin{figure*}[t]
\centering
\resizebox{\linewidth}{!}{
\noindent\fbox{%
    \parbox{21cm}{%
[\gray{this} morning$^{\blue{0.212}}$ I$^{\blue{0.145}}$ \gray{had} \gray{my} \gray{2nd} dose$^{\blue{0.028}}$ shot$^{\blue{0.075}}$.]$^{\blue{0.074}}$ [\hl{finally$^{\blue{0.054}}$, I$^{\blue{0.145}}$}\gray{\hl{'m}}\hl{ glad$^{\blue{0.865}}$ I$^{\blue{0.145}}$}\gray{\hl{am}}\hl{ fully$^{\blue{0.142}}$ vaccinated$^{\blue{0.215}}$}\gray{\hl{ 
now}}\gray{\hl{ and that}}]$^{\blue{0.463}}$ [\hl{I$^{\blue{0.145}}$}\gray{\hl{ will}} \hl{reach$^{\blue{0.206}}$}\hl{ 
full}\hl{ protection$^{\blue{0.531}}$}\gray{\hl{ in}}\gray{\hl{ about}}\gray{\hl{ two}}\hl{ weeks$^{\blue{0.105}}$}\gray{\hl{, but}}\gray{\hl{ idk}}]$^{\blue{0.217}}$ [\gray{it} feels$^{\blue{0.153}}$ like$^{\blue{0.363}}$ \gray{it} barely$^{\blue{0.06}}$ changes$^{\blue{0.095}}$ \gray{anything}]$^{\blue{0.187}}$ [\gray{I} \gray{was} waiting$^{\blue{0.121}}$ \gray{for} \gray{this} \gray{so} long$^{\blue{0.143}}$, \gray{and} \gray{now} I$^{\blue{0.145}}$ \gray{'m} \gray{just}... mildly$^{\blue{0.185}}$ disappointed$^{\blue{0.003}}$.]$^{\blue{0.093}}$  [right$^{\blue{0.174}}$ \gray{now} situation$^{\blue{0.118}}$ \gray{in} \gray{my} country$^{\blue{0.121}}$ \gray{is} more$^{\blue{0.147}}$ \gray{or} less$^{\blue{0.105}}$ okay$^{\blue{0.163}}$]$^{\blue{0.231}}$, [\gray{but} I$^{\blue{0.145}}$ \gray{am} \gray{still} WFH$^{\blue{0.164}}$]$^{\blue{0.251}}$ [\gray{and} \gray{my} office$^{\blue{0.121}}$ \gray{sent} \gray{us} official$^{\blue{0.142}}$ info$^{\blue{0.111}}$ \gray{that}]$^{\blue{0.142}}$ [we$^{\blue{0.131}}$ \gray{can} start$^{\blue{0.152}}$ coming$^{\blue{0.123}}$ \gray{back} \gray{to} \gray{the} office$^{\blue{0.121}}$ \gray{only} \gray{since} September$^{\blue{0.118}}$]$^{\blue{0.246}}$ [\gray{so} I$^{\blue{0.145}}$ \gray{'m} \gray{still} \gray{gonna} \gray{be} stuck$^{\blue{0.054}}$ \gray{at} home$^{\blue{0.104}}$ \gray{over} \gray{the} summer$^{\blue{0.131}}$]$^{\blue{0.076}}$. [\gray{then} \gray{also} I$^{\blue{0.145}}$ \gray{am} worried$^{\blue{0.074}}$ \gray{that} September$^{\blue{0.118}}$ return$^{\blue{0.142}}$ \gray{to} \gray{the} office$^{\blue{0.121}}$ \gray{won't} happen$^{\blue{0.103}}$]$^{\blue{0.054}}$  [\gray{because} everyone$^{\blue{0.131}}$ \gray{is} scaring$^{\blue{0.052}}$ \gray{is} \gray{that} \gray{we} \gray{will} \gray{have} another$^{\blue{0.117}}$ wave$^{\blue{0.186}}$ \gray{in} \gray{the} autumn$^{\blue{0.128}}$]$^{\blue{0.095}}$ . [\gray{so} literally$^{\blue{0.118}}$ \gray{what} \gray{is} \gray{the} point$^{\blue{0.142}}$]$^{\blue{0.131}}$

\vspace{0.1cm}

\hrule

\vspace{0.1cm}

[\gray{this} morning$^{\red{0.196}}$ I$^{\red{0.132}}$ \gray{had} \gray{my} \gray{2nd} dose$^{\red{0.073}}$ shot$^{\red{0.113}}$.]$^{\red{0.136}}$ [finally$^{\red{0.051}}$, I$^{\red{0.132}}$ \gray{'m} glad$^{\red{0.004}}$ I$^{\red{0.132}}$ \gray{am} fully$^{\red{0.094}}$ vaccinated$^{\red{0.108}}$ \gray{now} \gray{and} \gray{that}]$^{\red{0.055}}$ [I$^{\red{0.132}}$ \gray{will} reach$^{\red{0.141}}$ \gray{full} protection$^{\red{0.076}}$ \gray{in} \gray{about} \gray{two} weeks$^{\red{0.103}}$. \gray{but} \gray{idk},]$^{\red{0.134}}$  [\gray{it} feels$^{\red{0.125}}$ like$^{\red{0.100}}$ \gray{it} barely$^{\red{0.315}}$ changes$^{\red{0.169}}$ anything$^{\red{0.147}}$]$^{\red{0.148}}$  [I$^{\red{0.132}}$ \gray{was} waiting$^{\red{0.196}}$ \gray{for} \gray{this} \gray{so} \gray{long}, \gray{and} \gray{now} I$^{\red{0.132}}$ \gray{'m} \gray{just}... mildly$^{\red{0.075}}$ disappointed$^{\red{0.241}}$]$^{\red{0.185}}$ 
[right$^{\red{0.105}}$ \gray{now} situation$^{\red{0.143}}$ \gray{in} \gray{my} country$^{\red{0.105}}$ \gray{is} more$^{\red{0.122}}$ \gray{or} less$^{\red{0.162}}$ okay$^{\red{0.104}}$]$^{\red{0.142}}$, [\gray{but} I$^{\red{132}}$ \gray{am} \gray{still} WFH$^{\red{0.151}}$]$^{\red{0.152}}$ [\gray{and} \gray{my} office$^{\red{0.117}}$ \gray{sent} \gray{us} official$^{\red{0.123}}$ info$^{\red{0.113}}$ \gray{that}]$^{\red{0.105}}$ [\gray{we} \gray{can} start$^{\red{0.106}}$ coming$^{\red{0.121}}$ \gray{back} \gray{to} \gray{the} office$^{\red{0.123}}$ \gray{only} \gray{since} September$^{\red{0.152}}$.]$^{\red{0.142}}$  [\gray{\hl{so}}\hl{ I$^{\red{0.132}}$}\gray{\hl{ 'm}}\gray{\hl{ still}}\gray{\hl{ gonna}}\gray{\hl{ be}}\hl{ stuck$^{\red{0.386}}$}\gray{\hl{ at}}\hl{ home$^{\red{0.164}}$}\gray{\hl{ over}}\gray{\hl{ the}}\hl{ summer$^{\red{0.126}}$}]$^{\red{0.216}}$. [\gray{\hl{then}}\gray{\hl{ also}}\hl{ I$^{\red{0.132}}$}\gray{\hl{ am}} \hl{worried$^{\red{0.523}}$}\gray{\hl{ that}}\hl{ September$^{\red{0.152}}$}\hl{ return$^{\red{0.107}}$}\gray{\hl{ to}}\gray{\hl{ the}}\hl{ office$^{\red{0.111}}$}\gray{\hl{ won't}}\hl{ happen$^{\red{0.142}}$}]$^{\red{0.599}}$ [\gray{\hl{because}}\hl{ everyone$^{\red{0.116}}$}\gray{\hl{ is}}\hl{ scaring$^{\red{0.387}}$}\gray{\hl{ is}}\gray{\hl{ that}}\gray{\hl{ we}}\gray{\hl{ will}}\gray{\hl{ have}}\hl{ another$^{\red{0.175}}$}\hl{wave$^{\red{0.221}}$}\gray{\hl{ in}}\gray{\hl{ the}}\hl{ autumn$^{\red{0.131}}$}]$^{\red{0.372}}$. [\gray{so} literally$^{\red{0.101}}$ \gray{what} \gray{is} \gray{the} point$^{\red{0.121}}$]$^{\red{0.155}}$
    }%
}
}

\caption{Word-level Emotion-Aware PageRank scores and sentence-level meaning scores for the \textcolor{blue}{joy} (Upper Box) and \textcolor{red}{fear} (Lower Box) emotions. The term relevance score is superscripted to each word (i.e., $w^{score}$), while the meaning score of sentences is superscripted at the end of the sentence (i.e., $[.]^{score}$). Gold summaries are \hl{highlighted}.}
\vspace{-4mm}
\label{anecdotal_example}
\end{figure*}

\paragraph{Experimental Setup.} We carry out our experiments on an Nvidia A$5000$ GPU. We use the HuggingFace Transformers \cite{DBLP:journals/corr/abs-1910-03771} library for our Sentence-BERT implementation and we will make the code for our methods and data available for reasearch purposes. We report the performance in terms of Rouge-2 and Rouge-L \cite{lin-2004-rouge} to evaluate the summarization performance. Additionally, we also calculate the performance in terms of F1 and show the results in Appendix \ref{f1_results}. We provide extensive details about the hyperparameters used in EAP and the baselines, such as our various thresholds and constants in Appendix \secref{hyperparameters}.

% \jl{Don't forget parameters and performances of the bert emotion classifiers too.}
% \cornelia{how is the F1 calculated}

\noindent
\paragraph{Results.} We show the results obtained in Table \ref{tab:main_table}. First, we note that emotion-specific approaches outperform the emotion-oblivious methods considerably. Notably, EmoIntensity outperforms PacSum by an average of $1.1\%$ in Rouge-2. Among the emotion-specific baselines,
% \jl{what are ``weak'' baselines?? I think it's better to just compare with two groups: (1) emotion-agnostic; (2) emotion-specific. Within each you can break it down by how close it is to the final model.} 
EmoIntensity, which uses the intensity of emotion words to extract relevant sentences for a particular emotion obtains good performance, outperforming the EmoLex method by $5.1\%$ Rouge-2 on disgust and $3.3\%$ on fear. This result emphasizes that having a degree of association between a word and an emotion (i.e., the intensity) is a stronger signal than the plain word-emotion association in our emotion-based extractive summarization context.

\textbf{EAP consistently yields the highest results both in terms of Rouge-2 and Rouge-L} compared to the other approaches. Concretely, we obtain an average improvement of $2.7\%$ in Rouge-L and $2.5\%$ in Rouge-2 score over our strongest EmoIntensity baseline. For example, on anger and joy we see improvements in Rouge-2 of $1.7\%$ and $6.3\%$ respectively. Moreover, our emotion-aware PageRank considerably outperforms TextRank \cite{mihalcea-tarau-2004-textrank} by as much as $5.5\%$ Rouge-L and $4.5\%$ Rouge-2 on average.

\paragraph{Emotion Detection.}
While EAP shows strong results in our emotion trigger summarization experiments, we want to evaluate our approach in a traditional emotion detection task. To this end, we ask how well EAP can detect emotions at the post level. Given a post P, we label the post with emotion $e$ if we identify any sentence $s \in P$ as a summary for $e$. If no sentence is selected to be included in the summary, we consider that EAP does not predict $e$. 

We show the results obtained in Table \ref{tab:emotion-detection}, where we compare EAP to lexical methods (EmoLex and EmoIntensity) and a domain adaptation method, which trains a BERT \cite{devlin-etal-2019-bert} model on the GoEmotions dataset \cite{demszky-etal-2020-goemotions}. We observe that EAP consistently outperforms prior work on all the emotions by an average of $0.9\%$ in F1 score. Notably, we see $1.5\%$ improvements in F1 on fear and $1.9\%$ on anticipation.

\paragraph{Ablation Study.} We perform a thorough ablation study to tease apart and analyze the components lead to the success of EAP. First, we analyze the influence of emotion intensity on the performance of the model. Here, we slightly modify the importance function from Equation \ref{importance_function} to a constant value. Instead of using the variable $int_{e}(w)$ we use a constant value $c^{e}$ where $c^{e} > c$. Intuitively, we still bias the model towards a particular emotion $e$, however, every word associated with $e$ weighs equal in this ablated version of EAP. We denote this modification of the algorithm by \emph{-int}. Second, we remove the $meaning$ score $M_{e}$ from our algorithm and use only the word-based relevance $\mathcal{R}_{e}$. This approach is denoted by \emph{-sim}. We also analyze the behaviour of EAP when removing both components.

We show the results obtained in Table \ref{tab:ablation}. Removing emotion intensity leads to a performance degradation of $1\%$ in Rouge-L while the lack of our similarity module decreases the performance by $1.2\%$ in Rouge-L. Removing both further decreases the performance by $2.9\%$ in Rouge-2. These results emphasize that both similarity and intensity are core components of EAP and both consistently contribute to its success.

\looseness=-1
\paragraph{Anecdotal Evidence.} 
To offer additional insights into our EAP, we provide anecdotal evidence in Figure \ref{anecdotal_example}, where we show a post expressing both joy and fear. We indicate for each word both its relevance for joy and for fear. Additionally, we show the meaning score for each sentence and emotion. Interestingly, we observe that the scores produced by our model are very relevant. For instance, \emph{protection} has a very large value for joy of $0.531$ and a very small value of $0.076$ for \emph{fear}. Along the same lines, \emph{worried} has a relevance of $0.523$ for \emph{fear} and $0.074$ for joy. The similarity scores are also accurate. For example, \emph{glad I am fully vaccinated} has a score for joy of $0.463$, $9$ times as large of the score of the same sentence for fear. We show additional analysis on the effect of the most relevant terms on EAP performance in Appendix \secref{model_analysis}.

% \subsection{Supervised Baselines}
% \tibi{this will be moved in Appendix -> benchmark dataset will link to this section since }

\section{Conclusion}
\looseness=-1
We introduce \dataset{}, a new benchmark dataset composed of $1,883$ Reddit posts annotated for the task emotion detection and extractive trigger summarization in the context of the COVID-19 pandemic. Our proposed Emotion-Aware Pagerank approach yields strong results on our datasets, consistently outperforming prior work in an unsupervised learning context. In the future, we plan to study abstractive trigger summarization from an unsupervised point of view to bridge the gap between the extractive and abstractive summarization performance. 

\section*{Limitations}
Since our EAP builds its graph representation from social media data, our method may carry inductive biases rooted in this type of data. Moreover, note that the scope of our study is limited to English social media posts and our approach does not consider inputs larger than $512$ tokens. Therefore using our approach in long document summarization may be challenging. Finally, the general applicability of EAP in a different domain is highly dependent on the existence of high-quality lexicons for the domain in question, which may not be available.

\section*{Acknowledgements}
This research was partially supported by National Science Foundation (NSF) grants IIS-1912887, IIS-2107487, ITE-2137846, IIS-2145479, IIS-2107524, IIS-2107487. We thank Jamie Pennebaker for useful discussions and comments. We also thank our reviewers for their insightful feedback and comments.

\bibliography{custom}
\bibliographystyle{acl_natbib}

\clearpage
\appendix

\section{\textsc{CovidET}\footnote{\url{https://github.com/honglizhan/CovidET}}}\label{covid-et}

\citet{zhan-etal-2022-feel} was the first to introduce the combined labeling of both emotions and (abstractive) summaries of their triggers on the domain of spontaneous speech (i.e., Reddit posts). They presented \CovidET{}, a corpus of $1,883$ Reddit posts manually annotated with $7$ emotions (namely \textit{anger}, \textit{anticipation}, \textit{joy}, \textit{trust}, \textit{fear}, \textit{sadness}, and \textit{disgust}) as well as abstractive summaries of the emotion triggers described in the post. The posts are curated from \texttt{r/COVID19\_support}\footnote{\url{https://www.reddit.com/r/COVID19_support/}}, a sub-Reddit for people seeking community support during COVID-19. To ensure the diversity of the data distribution, \CovidET{} consists of Reddit posts from two different timelines (before and during the Omicron variant). The posts in \CovidET{} are lengthy and emotionally rich, with an average of $156.4$ tokens and $2.46$ emotions per post. \CovidET{} serves as an ideal dataset to spur further research on capturing triggers of emotions in long social media posts.

%Their English dataset not only considers longer documents and multiple emotions, but also takes into account implicit emotions that are not expressed lexically and require reasoning.
Nevertheless, the combined labeling of emotions and free-form abstractive summarization of their triggers is difficult and time-consuming as it requires annotators to comprehend the document in depth. This fails to meet the time-sensitivity requirement in the face of major crises like COVID-19. Our work instead proposes to generate an extractive summarization of emotion triggers and studies the task of emotion detection and trigger summarization from an unsupervised learning perspective, which is robust to domain variations and beneficial in boosting understanding in time-critical periods.

\section{Annotation Scheme of \dataset{}}\label{appendix:annotation-scheme}
\begin{figure}[t]
    \centering
    \includegraphics[width=\columnwidth]{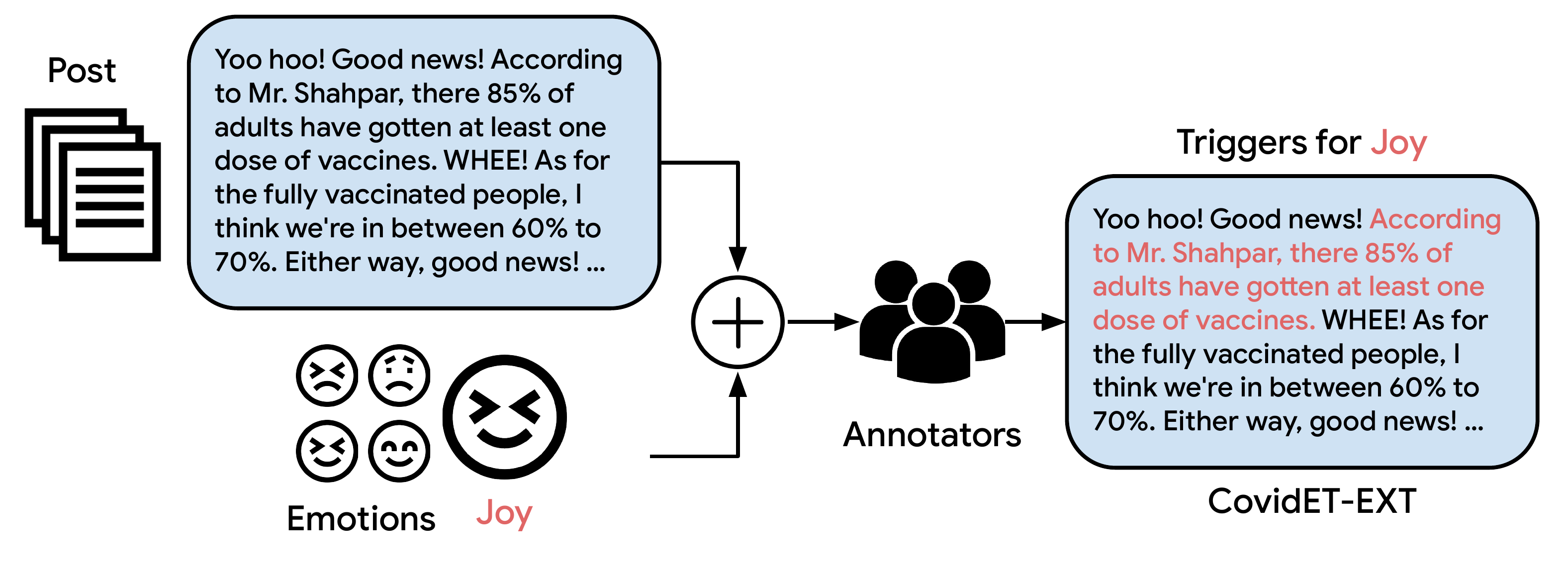}
    \caption{The process of collecting annotations for \dataset{}. The provided posts and annotated emotions are gathered from \CovidET{} \cite{zhan-etal-2022-feel}.}
    \label{fig:annotation-flow-chart}
\end{figure}

The process of collecting annotations for \dataset{} is shown in Figure \ref{fig:annotation-flow-chart}. Given a post and its annotations containing emotion $e$ from \CovidET{}, we ask annotators to highlight sentences in the post that best describe the trigger for emotion $e$. Rather than selecting text that expresses the emotion itself, we specifically instruct annotators to extract the events and how people make sense of the events that lead to the expression of the emotion. We use detailed examples provided by \citet{zhan-etal-2022-feel} to help our annotators better interpret the definition of emotion triggers.%\ts{references?examples?}

\section{Crowd Workers}
Both groups of annotators for \dataset{} come from the United States. The crowd workers are recruited from the Amazon Mechanical Turk crowdsourcing platform, with restrictions that their
locale is the US and that they have completed $500+$
HITs with an acceptance rate of at least $95\%$. The undergraduate students are hired from a university in the United States.

\section{Inter-annotator Agreement Among Undergraduate Students and Crowd Workers}\label{agreement}
As shown in Table \ref{tab:appendix-agreement}, the inter-annotator performance of the undergraduate students consistently exceeds the crowd workers.

\begin{table}[htbp]
  \centering
  \small
    \begin{tabular}{l|ccc}
          \toprule
          & \multicolumn{1}{c}{\textbf{Students}} & \multicolumn{1}{c}{\textbf{Crowd Workers}} \\
          \midrule
    \textbf{self-BLEU-2} & $0.466$ & $0.418$ \\
    \textbf{self-BLEU-3} & $0.456$ & $0.408$ \\
    \textbf{self-ROUGE-L} & $0.553$ & $0.489$ \\
        \bottomrule
    \end{tabular}
  \caption{Inter-annotator agreement among undergraduate students and crowd workers in \dataset{}.}
  \label{tab:appendix-agreement}
\end{table}

\section{Additional Analyses of \dataset{}}\label{appendix:data-analysis}

\begin{figure}[t]
    \centering
    \includegraphics[width=\columnwidth]{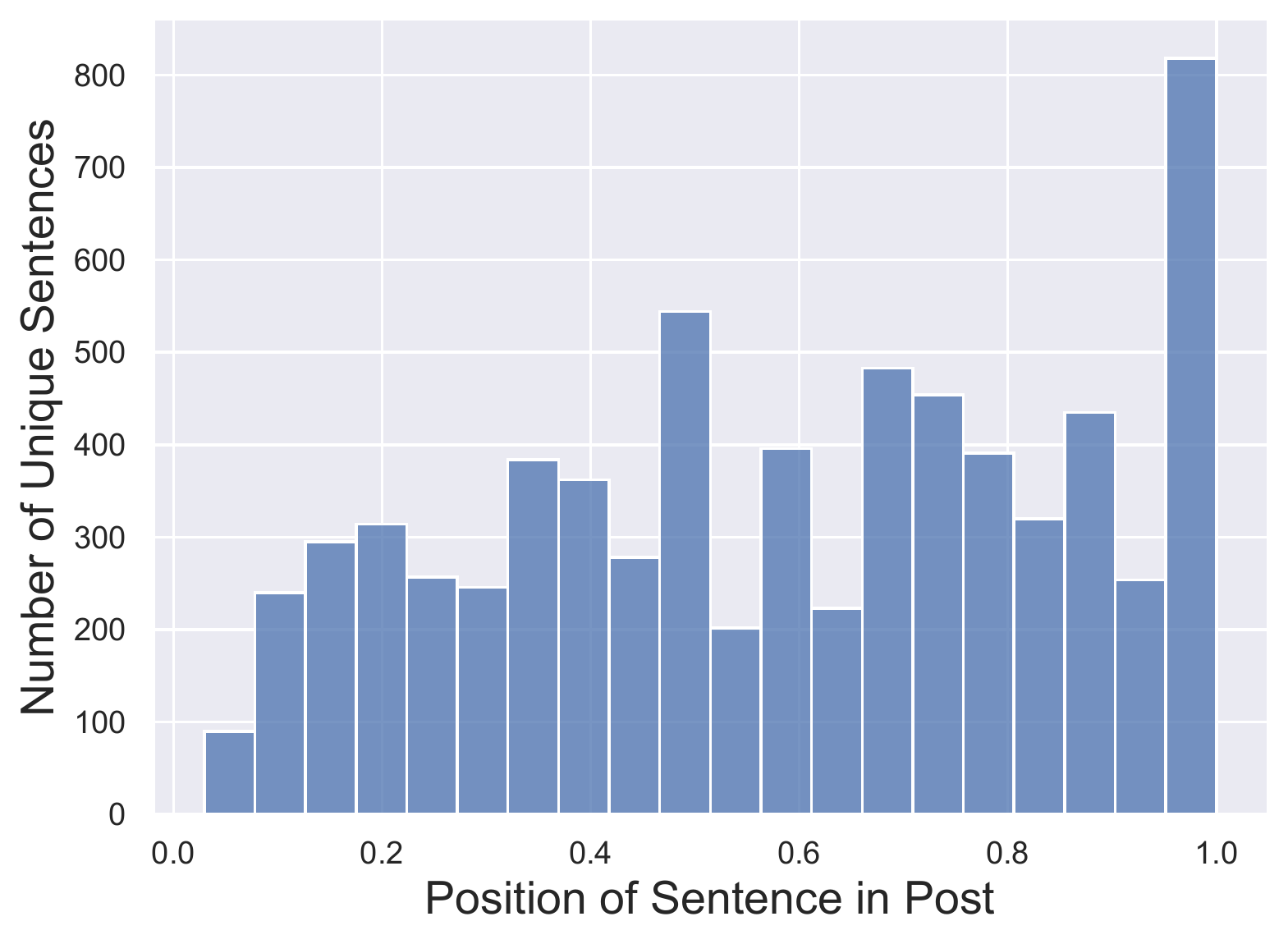}
    \caption{The sentence-level distribution of triggers in the original posts of \dataset{}.}
    \label{fig:ext-trigger-position}
\end{figure}

\begin{figure}[ht]
    \centering
    \includegraphics[width=\columnwidth]{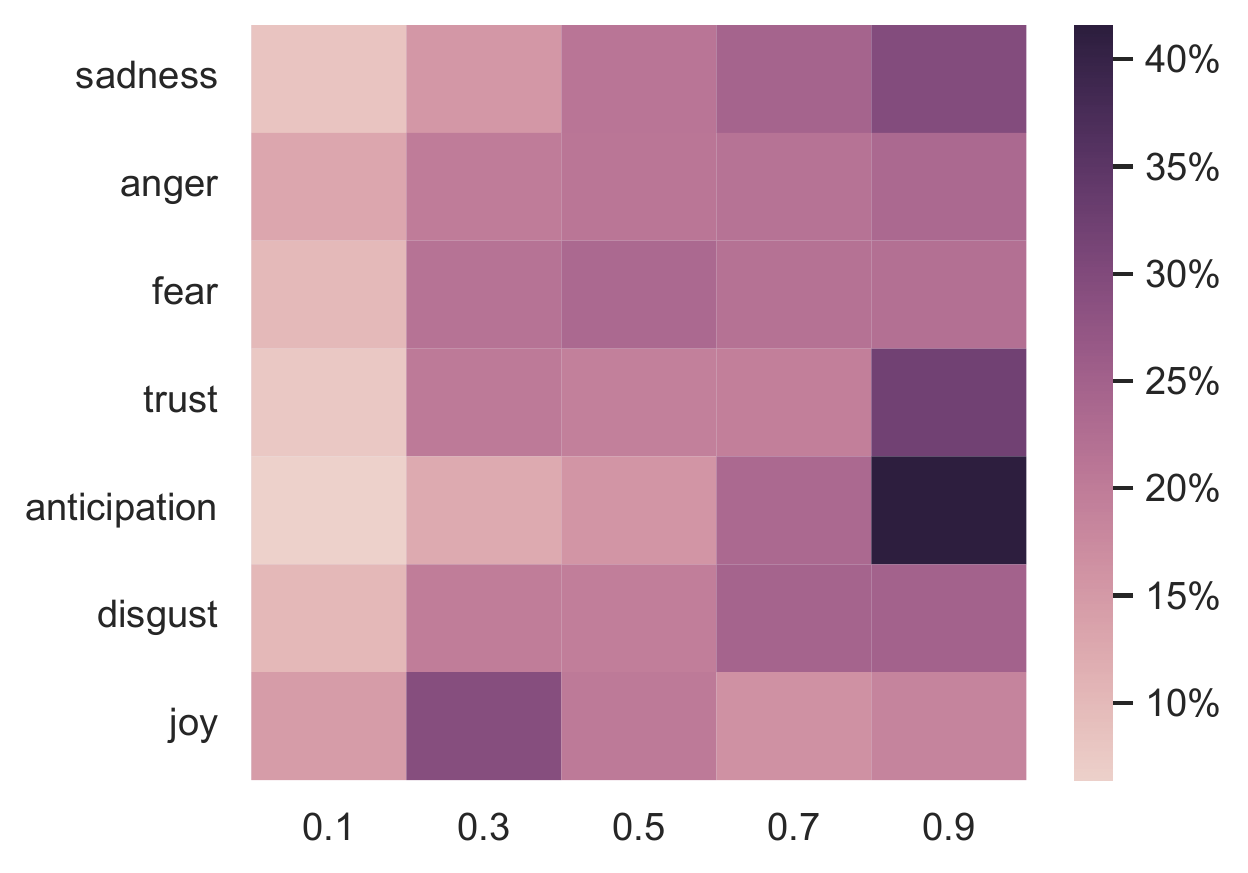}
    \caption{Heatmap of the distribution of triggers in the original posts of \dataset{}. The X-axis stands for the position of trigger sentences in the original post, and the colorbar exhibits the percentage of trigger sentences under the emotion label.}
    \label{fig:ext-trigger-heatmap}
\end{figure}

\paragraph{Trigger Positions.} We examine the position of the emotion trigger sentences in the original posts. The sentence-level distribution of the annotated triggers is reported in Figure \ref{fig:ext-trigger-position}. Results reveal that the trigger sentences spread evenly across the posts, with a large number of triggers clustering in the later parts of the post. This means that the emotion-trigger extractive summarization task is \emph{not} lead-based, unlike generic news summarization~\cite{fabbri2021summeval,10.1007/978-981-10-8639-7_25}. This is especially true for \textit{anticipation}, as demonstrated in Figure \ref{fig:ext-trigger-heatmap}.% \hz{done}

\begin{figure}[t]
    \centering
    \includegraphics[width=\columnwidth]{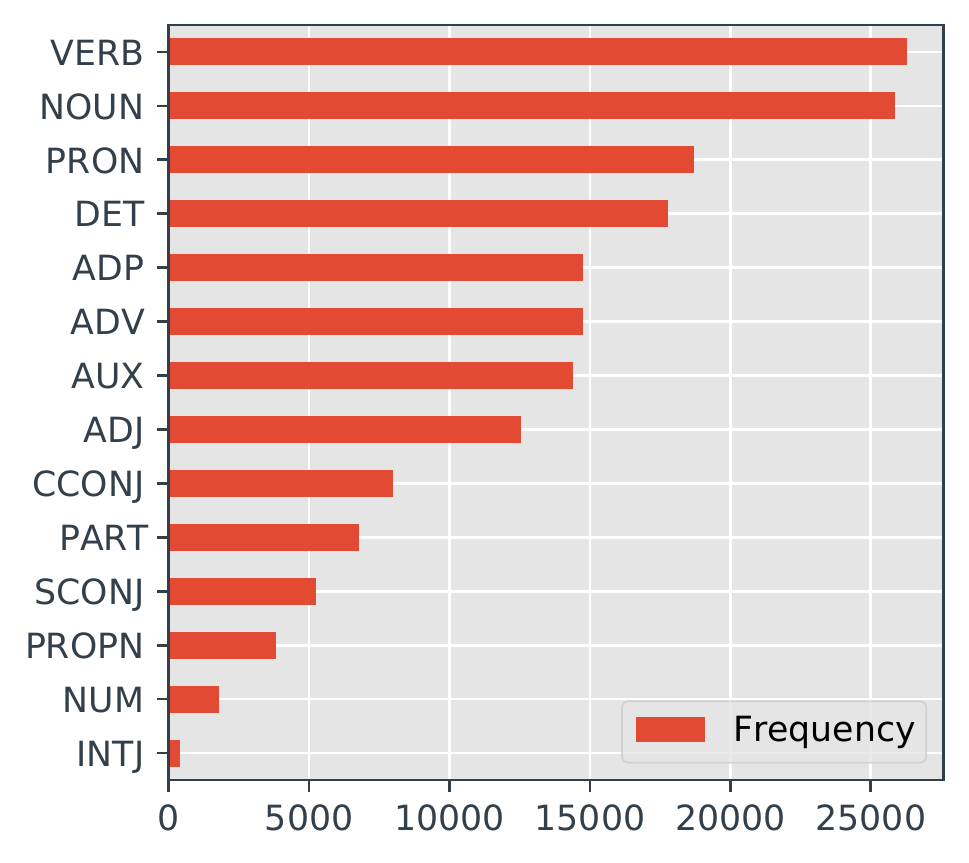}
    \caption{POS frequency distribution in \dataset{}.}
    \label{fig:ext-pos}
\end{figure}

\paragraph{Trigger Components.} In addition to the explicitness of emotion triggers, we also examine the syntactic components of the extractive summaries of emotion triggers. Results are shown in Figure \ref{fig:ext-pos}. We observe that nouns and verbs take up the majority of triggers, closely followed by the use of pronouns.

\paragraph{Pronoun Distributions.} Psycho-linguistic studies reveal that the analysis of function words such as pronouns can disclose psychological effects of life experiences and social processes \cite{campbell2003secret,tausczik2010psychological,Pennebaker2014WhenSW, pennebaker-2021-breakup,singh2018footwear}. Specifically, overusing the first-person singular pronouns may imply a high level of self-involvement, whereas the increased use of other pronouns may signify improvement of social engagement \cite{cohn2004linguistic,simmons2008hostile,kumari2017parallelization}.

\begin{figure*}[htpb]
    \centering
    \includegraphics[width=0.8\textwidth]{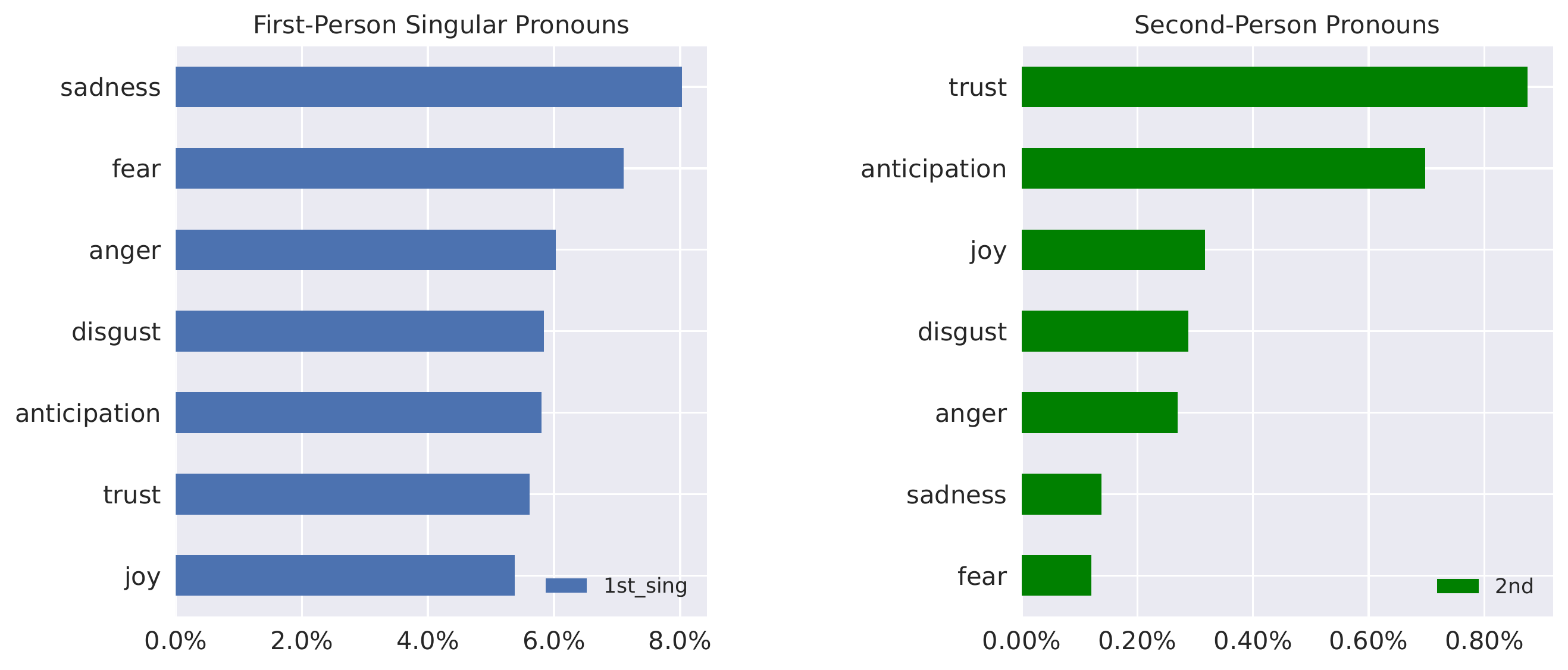}
    \caption{Average percentage of first-person singular and second-person pronouns in the annotated extractive summaries of emotion triggers of \dataset{}.}
    \label{fig:ext-pronouns}
\end{figure*}

We evaluate the percentage of personal pronoun usage per annotated emotion trigger sentence. In particular, we discover an inverse correlation between first-person singular pronouns (e.g., \textit{I}, \textit{me}, \textit{my}, \textit{mine}, \textit{myself}) and second-person pronouns (e.g., \textit{you}, \textit{your}, \textit{yours}, \textit{yourself}, \textit{yourselves}). We provide the average percentage of the personal pronouns per emotion trigger in Figure \ref{fig:ext-pronouns}. Further statistical tests reveal negative Pearson correlations between the percentage distribution of first-person singular pronouns and second-person pronouns in each emotion (with substantial significance in all $7$ emotions; shown in Table \ref{tab:stats-prons}). We note that when expressing negative emotions such as \textit{sadness} and \textit{fear}, authors used more first-person singular pronouns in triggers. On the other hand, authors used more second-person pronouns in expressing the triggers for positive emotions like \textit{joy} and \textit{trust}. The inverse correlation between first-person singular pronouns and second-person pronouns suggests more self-involvement in negative emotions and more social engagement in positive emotions in \dataset{}.

%2nd person pronouns: social orientation

\begin{table}[htbp]
  \small
  \centering
    \begin{tabular}{l||cl}
    \toprule
          & \multicolumn{1}{c}{Pearson's $r$} & \multicolumn{1}{c}{\textit{p}} \\
    \midrule
    anger & $-0.1288$ & $4.77e^{-06}$* \\
    fear  & $-0.0903$ & $1.45e^{-05}$* \\
    anticipation & $-0.1671$ & $1.13e^{-14}$* \\
    joy   & $-0.1634$ & $1.22e^{-03}$* \\
    sadness & $-0.0945$ & $2.05e^{-03}$* \\
    trust & $-0.1873$ & $7.74e^{-04}$* \\
    disgust & $-0.1167$ & $1.90e^{-02}$* \\
    \bottomrule
    \end{tabular}
  \caption{Pearson Correlation Coefficients between the percentage distribution of first-person singular pronouns and second-person pronouns among emotions in \dataset{}. * indicates $p$ value $< 0.05$.}
  \label{tab:stats-prons}
\end{table}

\paragraph{Topical Variations.} 
% \hz{@Tibi: any connections to model??}\jl{Move to appendix, and refer to it in the relevant discussions in the modeling section.} 
To better interpret the annotated emotion triggers, we train a multi-class bag-of-words logistic regression model to predict the emotion label of each annotated extractive emotion trigger sentence. The trained model's weights pertaining to each class of emotions are then extracted to locate the tokens that are most indicative of each emotion. The multi-class logistic regression model achieved a micro F1 score of $0.33$ after training and evaluating on our benchmark dataset. The most indicative tokens associated with each emotion are reported in Table \ref{tab:lr-weights}.

\paragraph{Connections to \CovidET{}.}
To understand the ties between \dataset{} and \CovidET{}, we measure the self-BERTScore between the extractive summaries of triggers from \dataset{} and the abstraction summaries of triggers from \CovidET{}. Results reveal that the average BERTScore F1 is $0.872$ between the extractive and abstractive summaries, indicating strong correlations between the two datasets.

\begin{figure}[t]
    \centering
    \resizebox{\linewidth}{!}{
    \includegraphics[width=\textwidth]{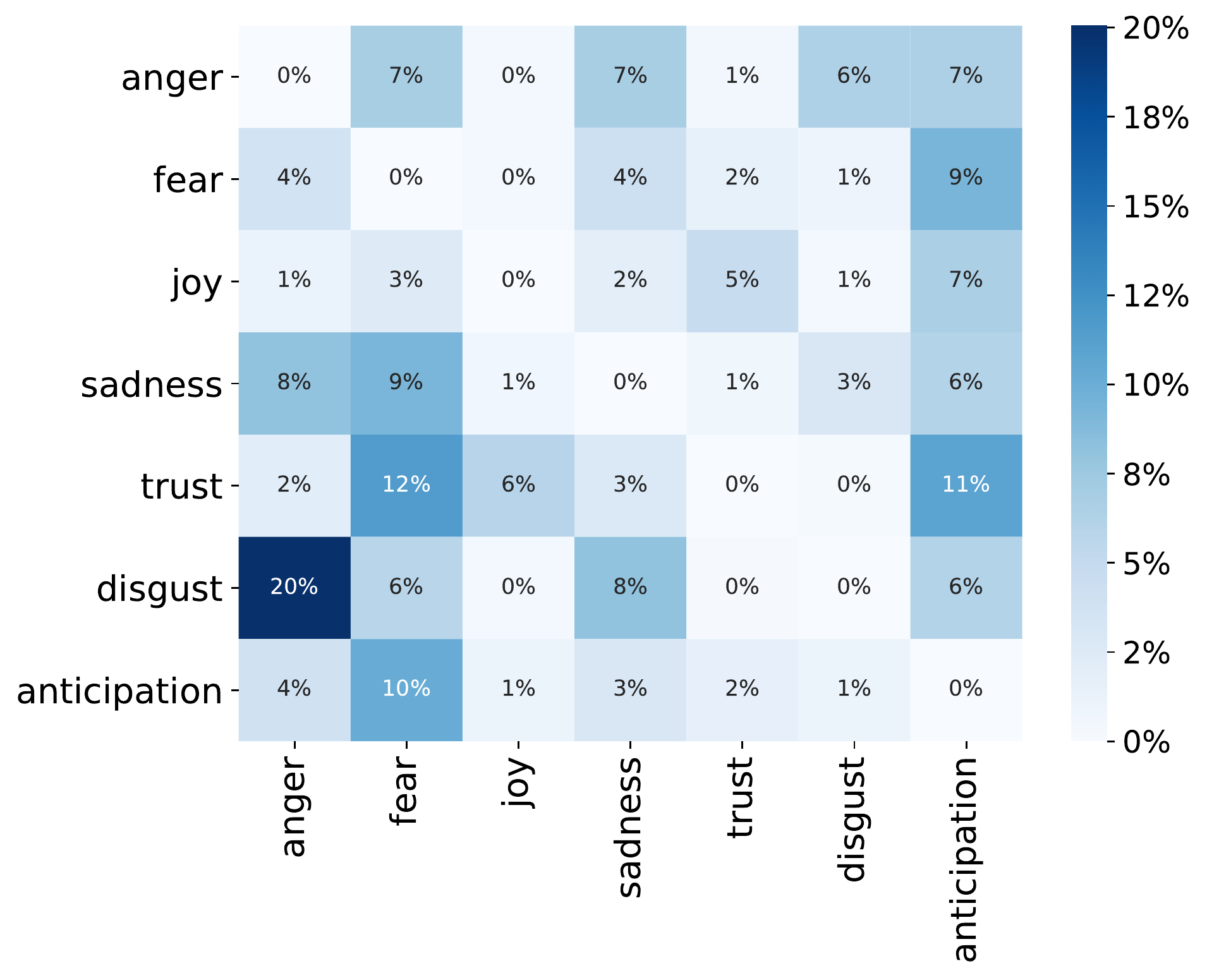}
    }
    \caption{Overlapping trigger sentences for different emotions. Cell $(i, j)$ represents the percentage of sentences that are triggers for emotion $i$ that are also triggers for emotion $j$ in \dataset{}.}
    \label{fig:trigger-heatmap}
\end{figure}

\paragraph{Same Triggers for Different Emotions.} The status of overlapping trigger sentences for different emotions is shown in Figure \ref{fig:trigger-heatmap}. Specifically, we measure the percentage of sentences that are triggers for an emotion $i$ that are also triggers for emotion $j$ in \dataset{}.

\begin{table*}[htpb]
  \centering
  \setlength{\tabcolsep}{3pt}
  \small
    \resizebox{16cm}{!}{
    \begin{tabular}{r|cc|cc|cc|cc|cc|cc|cc}
          & \multicolumn{2}{c}{ANGER} & \multicolumn{2}{c}{DISGUST} & \multicolumn{2}{c}{FEAR} & \multicolumn{2}{c}{JOY} & \multicolumn{2}{c}{SADNESS} & \multicolumn{2}{c}{TRUST} & \multicolumn{2}{c}{ANTICIPATION} \\
          & Token & Weight & Token & Weight & Token & Weight & Token & Weight & Token & Weight & Token & Weight & Token & Weight \\
          \toprule
    0     & annoying & 7.08  & prodded & 5.33  & grandpa & 7.11  & grateful & 6.84  & unrelated & 5.93  & reacting & 6.19  & wreck & 5.69 \\
    1     & upset & 6.42  & guard & 5.05  & shd   & 6.92  & thankfully & 5.89  & believing & 5.93  & okay  & 5.34  & monger & 5.53 \\
    2     & angry & 5.81  & maskless & 4.94  & freaking & 6.12  & happy & 5.43  & sad   & 5.80  & ineffective & 5.12  & harm  & 5.44 \\
    3     & ridiculed & 5.43  & wiped & 4.87  & expose & 6.09  & fantastic & 4.76  & devastated & 5.22  & stock & 5.06  & statistic & 5.43 \\
    4     & milder & 5.32  & care  & 4.84  & pass  & 6.07  & glad  & 4.72  & fault & 5.09  & cheer & 5.00  & question & 5.28 \\
    5     & ideal & 5.25  & nicely & 4.75  & afraid & 5.69  & provide & 4.65  & antivax & 5.08  & haven & 5.00  & waited & 5.04 \\
    6     & realized & 5.20  & experiencing & 4.44  & fear  & 5.60  & ended & 4.52  & depression & 5.00  & accepted & 4.55  & infectious & 5.04 \\
    7     & strongly & 5.20  & beginning & 4.41  & pills & 5.55  & million & 4.19  & disappear & 4.93  & affirmed & 4.26  & questioning & 4.97 \\
    8     & wtf   & 5.18  & coronavirus & 4.40  & scared & 5.52  & success & 4.09  & dead  & 4.81  & psychiatrist & 4.11  & june  & 4.86 \\
    9     & centered & 5.08  & dumbass & 4.39  & venting & 5.49  & effective & 3.89  & virtually & 4.77  & worried & 4.04  & wait  & 4.83 \\
    \bottomrule
    \end{tabular}}
  \caption{The tokens with the most positive weights for each emotion in a multi-class bag-of-words logistic regression model trained to classify the emotion indicated by the trigger sentences.}
  \label{tab:lr-weights}
\end{table*}

\begin{table*}[!htbp]
\setlength{\tabcolsep}{3pt}
\centering
\small
\resizebox{16cm}{!}{%
\begin{tabular}{r|cc|cc|cc|cc|cc|cc|cc|cc}
 & \multicolumn{2}{c}{\textsc{anger}} & \multicolumn{2}{c}{\textsc{disgust}} & \multicolumn{2}{c}{\textsc{fear}} & \multicolumn{2}{c}{\textsc{joy}} & \multicolumn{2}{c}{\textsc{sadness}} & \multicolumn{2}{c}{\textsc{trust}} & \multicolumn{2}{c}{\textsc{anticipation}} & \multicolumn{2}{c}{\textsc{avg}} \\
& R-2 & R-L & R-2 & R-L & R-2 & R-L & R-2 & R-L & R-2 & R-L & R-2 & R-L & R-2 & R-L & R-2 & R-L \\
 \toprule
\textsc{BART-FT-JOINT} & $0.335$ & $0.371$ & $0.299$ & $0.312$ & $0.377$ & $0.384$ & $0.304$ & $0.335$ & $0.375$ & $0.370$ & $0.254$ & $0.276$ & $0.333$ & $0.338$ & $0.325$ & $0.340$ \\ 
\textsc{BERT} & $0.329$ & $0.367$ & $0.291$ & $0.304$ & $0.372$ & $0.376$ & $0.293$ & $0.295$ & $0.361$ & $0.363$ & $0.242$ & $0.268$ & $0.323$ & $0.332$ & $0.315$ & $0.329$ \\ 
\midrule
\textsc{EAP} & $0.324$  &	$0.348$	& $0.285$ &	$0.296$	& $0.364$ & $0.373$& $0.285$ & $0.319$ & $0.348$ & $0.354$ & $0.239$ & $0.264$	& $0.319$ &$0.324$	& $0.309$ &$0.325$	 \\
\bottomrule
\end{tabular}
}
\caption{Comparison between EAP and supervised approaches.}
\label{tab:supervised-comparison}
\end{table*}

\section{Supervised Extractive Summarization}
\label{supervised-extractive-summarization}

Although our focus is exclusively on unsupervised approaches to eliminate the reliance on labeled data, we note that Covid-EXT can be a suitable benchmark for developing supervised methods as well. In this section, we compare two supervised methods against our unsupervised EAP. We experiment with two methods for emotion trigger extraction. \textbf{1)} First, we experiment with the BART-FT-JOINT \cite{zhan-etal-2022-feel} model which is trained to jointly predict emotions and their summary. We train this model on the training set of Covid-EXT in a supervised manner. Second we employ a simple \textbf{2)} BERT \cite{devlin-etal-2019-bert} classifier that is trained in a supervised manner to detect emotions at sentence level. We consider as positive examples the sentences that are included in the summary, and negative examples the rest of the sentences. Note that we train $7$ different models, one for each emotion. 

We show the results obtained in Table \ref{tab:supervised-comparison}. We observe that BART-FT-JOINT outperforms our EAP considerably by $1.5\%$ in Rouge-L score. However, we see that the BERT-based approach is much closer to the performance of the unuspervised EAP, outperforming it by less than $1\%$ in Rouge-L and F1.

% \begin{table}[!htbp]
% \setlength{\tabcolsep}{2pt}
% \centering
% \small
% \resizebox{\columnwidth}{!}{%
% \begin{tabular}{l|cccccccc}
%  & \textsc{ang} & \textsc{dsg} & \textsc{fer} & \textsc{joy} & \textsc{sdn} & \textsc{trt} & \textsc{anc} & \textsc{avg} \\
%  \toprule
% \textsc{HEm} & $0.507$ & $0.514$ & $0.548$ & $0.554$ & $0.496$ & $0.527$ & $0.573$ & $0.527$ \\ 
% \textsc{CEm} & $0.523$ & $0.521$ & $0.528$ & $0.569$ & $0.520$ & $0.567$ & $0.582$ & $0.527$ \\
% \midrule
% \textsc{EAP} & $\mathbf{0.593}$  & $\mathbf{0.595}$ & $\mathbf{0.583}$ & $\mathbf{0.649}$ &  $\mathbf{0.581}$  & $\mathbf{0.606}$  & $\mathbf{0.612}$ &  $\mathbf{0.593}$ \\
% \bottomrule
% \end{tabular}
% }
% \caption{Emotion detection results of our models in terms of F-1.}
% \label{tab:detection-cemo-hemo}
% \end{table}

% \section{Summarization and Emotion Detection using CancerEMO and HurricaneEMO}
% \label{summarization-cemo-hemo}
% We show the summarization and emotion detection results obtained using the HurricaneEMO \cite{desai-etal-2020-detecting} and CancerEMO \cite{sosea-caragea-2020-canceremo} datasets in Table \ref{tab:summarization-cemo-hemo} and \ref{tab:detection-cemo-hemo}.

\section{Hyperparameters}
\label{hyperparameters}
In this section we detail the values of the hyperparameters used and the search space considered in the development of our EAP. First in terms of the constant $c$ in Equation \ref{importance_function}, we experiment with values in the range $0.1 \rightarrow 0.5$ but observed that $0.1$ works well. We mentioned that the minimum frequency of a word necessary for selection in our vocabulary $V$ is $20$. We also experimented with other values ranging from $5$ to $50$. The threshold $t$ from Equation \ref{final_threshold} is emotion-specific and inferred using the validation set. We experiment with values between $0.2$ and $0.7$ and observed that $0.35$ works well in general.

\section{Model Analysis}
\label{model_analysis}
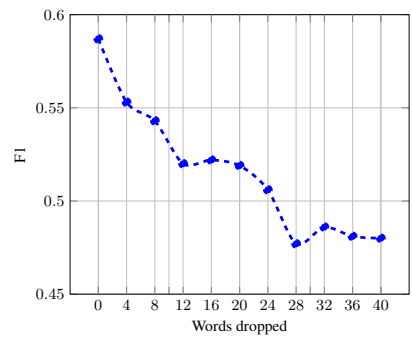
\begin{figure}[t]
\centering
\resizebox{.7\linewidth}{!}{
\begin{tikzpicture}
 \small
\begin{axis} [
ylabel={F1},
xlabel={Words dropped},
ymin=0.45, ymax=0.6,
grid,
xticklabels={},
extra x ticks={0,4,...,40},
every axis plot/.append style={ultra thick},
]
\addplot[smooth,mark=*,blue,dashed] plot coordinates {
    (0,0.587) 
    (4,0.553) 
    (8,0.543) 
    (12,0.520) 
    (16,0.522)
    (20,0.519)
    (24,0.506)
    (28,0.477)
    (32,0.486) 
    (36,0.481) 
    (40,0.480) 
};
\end{axis}
\end{tikzpicture}
}
\caption{Average F-1 obtained when dropping the top $k$ (with k from $0$ to $40$) highest relevance nodes in the graph.}
\label{analysis_dropped_terms}
\end{figure}

To offer additional insights into our approach, we show in Figure \ref{analysis_dropped_terms} an analysis on the effect of the top relevant terms on the performance of EAP. For each emotion, we experiment with completely dropping the top $k$ most relevant terms (i.e., words) in the graph, with $k$ ranging from $1$ to $40$ and report the average performance obtained. This analysis can be seen as a way to measure the reliance of EAP and the top relevant words.    We observe that the performance drops considerably while dropping the first $28$ terms and the starts to plateau.

\section{Extractive Summarization Results in terms of F1}
\label{f1_results}

In Table \ref{tab:main_table_f1_only} we present the performance on extractive summarization in terms of F1. While Rouge captures the overlap between extracted summaries and human references at word level, F1 measures the number of extracted sentences from the post that are correctly part of the gold summary (human references). Specifically, we compute F1 as if we dealt with a traditional classification problem. For every emotion, the sentences belonging to the trigger summaries are positive examples, and all the other sentences are negative examples. If our EAP model selects a sentence that does not appear in the trigger summary, we view it as a false positive. On the other hand, if our EAP model does not extract a sentence which belongs to the trigger summary, we count it as a false negative. We calculate F1 as the harmonic mean between precision and recall.

\begin{table*}[!htbp]
\setlength{\tabcolsep}{3pt}
\centering
\small
\begin{tabular}{r|c|c|c|c|c|c|c|c}
 & \textsc{anger} & \textsc{disgust} & \textsc{fear} & \textsc{joy} & \textsc{sadness} & \textsc{trust} & \textsc{anticipation} & \textsc{avg} \\
 \toprule
\textsc{1-sent}	&$0.14$	& $0.07$	& $0.159$ & 	$0.113$& 	$0.097$	&$0.197$&	$0.235$&	$0.144$ \\ 
\textsc{3-sent}	&$0.306$	&$0.182$	&$0.300$&	$0.275$	&$0.241$&	$0.270$&	$0.268$&	$0.263$ \\ 
\textsc{PacSum}	&$0.297$&	$0.179$	&$0.296$	&$0.280$&	$0.246$&	$0.271$&	$0.276$&	$0.263$	\\ 
\textsc{PreSumm}&	$0.302$&	$0.189$&	$0.302$&	$0.283$&	$0.241$&	$0.273$&	$0.274$	&$0.266$ \\
\textsc{TextRank}	&$0.286$&	$0.165$	&$0.289$	&$0.274$&	$0.239$&	$0.270$	&$0.211$	&$0.247$ \\
\midrule
\textsc{EmoLex}	&$0.238$	&$0.248$	&$0.320$	&$0.238$	&$0.298$&	$0.200$	&$0.218$&	$0.253$	 \\
\textsc{EmoIntensity}&	$0.298$&	$0.221$	&$0.347$	&$0.293$	&$0.325$	&$0.274$	&$0.272$	&$0.284$ \\
\textsc{BERT-GoEmo}&	$0.264$&	$0.215$&	$0.308$&	$0.216$&	$0.312$	&$0.201$	&$0.253$	&$0.269$ \\
\midrule
\textsc{eap}&	$\mathbf{0.315^{\dagger}}$	&$\mathbf{0.251^{\dagger}}$&	$\mathbf{0.361^{\dagger}}$	&$\mathbf{0.305^{\dagger}}$	&$\mathbf{0.354^{\dagger}}$&	$\mathbf{0.299^{\dagger}}$&	$\mathbf{0.285^{\dagger}}$&	$\mathbf{0.310^{\dagger}}$ \\
\bottomrule
\end{tabular}
\caption{Results of our models in terms of F1. We assert significance$^{\dagger}$ using a bootstrap test where we resample our dataset $50$ times with replacement (with a sample size of $500$) and $p<0.05$.}
\label{tab:main_table_f1_only}
\end{table*}

\end{document}